\documentclass[11pt]{article}

\usepackage[table]{xcolor} 
\usepackage{booktabs}      
\usepackage[preprint]{acl}

\usepackage{times}
\usepackage{latexsym}
\usepackage{booktabs}
\usepackage{multirow}
\usepackage{xcolor}
\usepackage{graphicx} 
\usepackage[absolute,overlay]{textpos}

\usepackage{subcaption} 

\usepackage{tabularx}

\usepackage{booktabs}
\usepackage{multirow}
\usepackage{array}
\usepackage{makecell}
\usepackage{amsmath}
\usepackage{amssymb}
\usepackage[ruled,vlined]{algorithm2e}

\usepackage{adjustbox} 
\usepackage{enumitem}
\usepackage{hyperref}
\usepackage[nameinlink,capitalize,noabbrev]{cleveref}
\usepackage{titletoc} 

\newcolumntype{L}[1]{>{\raggedright\arraybackslash}p{#1}}
\newcolumntype{C}[1]{>{\centering\arraybackslash}p{#1}}

\usepackage{tabularx}
\usepackage{makecell}
\usepackage{array}
\newcolumntype{Y}{>{\centering\arraybackslash}X} 

\usepackage[T1]{fontenc}

\usepackage[utf8]{inputenc}

\usepackage{microtype}

\usepackage{inconsolata}

\usepackage{graphicx}

\title{Evolving Beyond Snapshots: Harmonizing Structure and Sequence via Entity State Tuning for Temporal Knowledge Graph Forecasting}

\author{
\textbf{Siyuan Li\textsuperscript{1}}\thanks{Corresponding authors.}\thanks{Equal contribution.},
\textbf{Yunjia Wu\textsuperscript{1}}\footnotemark[2],
\textbf{Yiyong Xiao\textsuperscript{2}}\footnotemark[2],
\textbf{Pingyang Huang\textsuperscript{1}},
\\
\textbf{Peize Li\textsuperscript{3}},
\textbf{Ruitong Liu\textsuperscript{4}}\footnotemark[1],
\textbf{Yan Wen\textsuperscript{5}},
\textbf{Te Sun\textsuperscript{6}}
\\
\textsuperscript{1} Dalian University of Technology \quad
\textsuperscript{2} Shenzhen University of Advanced Technology \\
\textsuperscript{3} King's College London \quad
\textsuperscript{4} Peking University \\
\textsuperscript{5} Beijing Institute of Technology \quad
\textsuperscript{6} Shanghai Jiao Tong University
\\
yuanlsy@mail.dlut.edu.cn, ruitong.jerry@gmail.com
}

\begin{document}
\maketitle


\begin{abstract}

Temporal knowledge graph (TKG) forecasting requires predicting future facts by jointly modeling structural dependencies within each snapshot and temporal evolution across snapshots. However, most existing methods are stateless: they recompute entity representations at each timestamp from a limited query window, leading to episodic amnesia and rapid decay of long-term dependencies. To address this limitation, we propose Entity State Tuning (EST), an encoder-agnostic framework that endows TKG forecasters with persistent and continuously evolving entity states. EST maintains a global state buffer and progressively aligns structural evidence with sequential signals via a closed-loop design. Specifically, a topology-aware state perceiver first injects entity-state priors into structural encoding. Then, a unified temporal context module aggregates the state-enhanced events with a pluggable sequence backbone. Subsequently, a dual-track evolution mechanism writes the updated context back to the global entity state memory, balancing plasticity against stability. Experiments on multiple benchmarks show that EST consistently improves diverse backbones and achieves state-of-the-art performance, highlighting the importance of state persistence for long-horizon TKG forecasting.

\end{abstract}

\section{Introduction}

Temporal Knowledge Graphs (TKGs) extend static facts into quadruples $(s, r, o, t)$, organizing knowledge as a sequence of chronological snapshots~\cite{cai2022temporal,wang2023survey,liu2025neural,lin2026ontotkge}. This temporal grounding is pivotal for reasoning in dynamic environments, enabling applications ranging from Question Answering~\cite{wang2024kgquestion,liu2026joint} to Recommender Systems~\cite{wang2025knowledge}. 

Unlike static reasoning, TKG forecasting~\cite{liao2024gentkg} demands extrapolating future events by deciphering the complex interplay between two orthogonal information flows: the \textit{structural flow} (intra-snapshot topological dependencies) and the \textit{sequential flow} (inter-snapshot temporal evolution).

\begin{figure*}[t]
    \centering
    \includegraphics[width=0.97\linewidth]{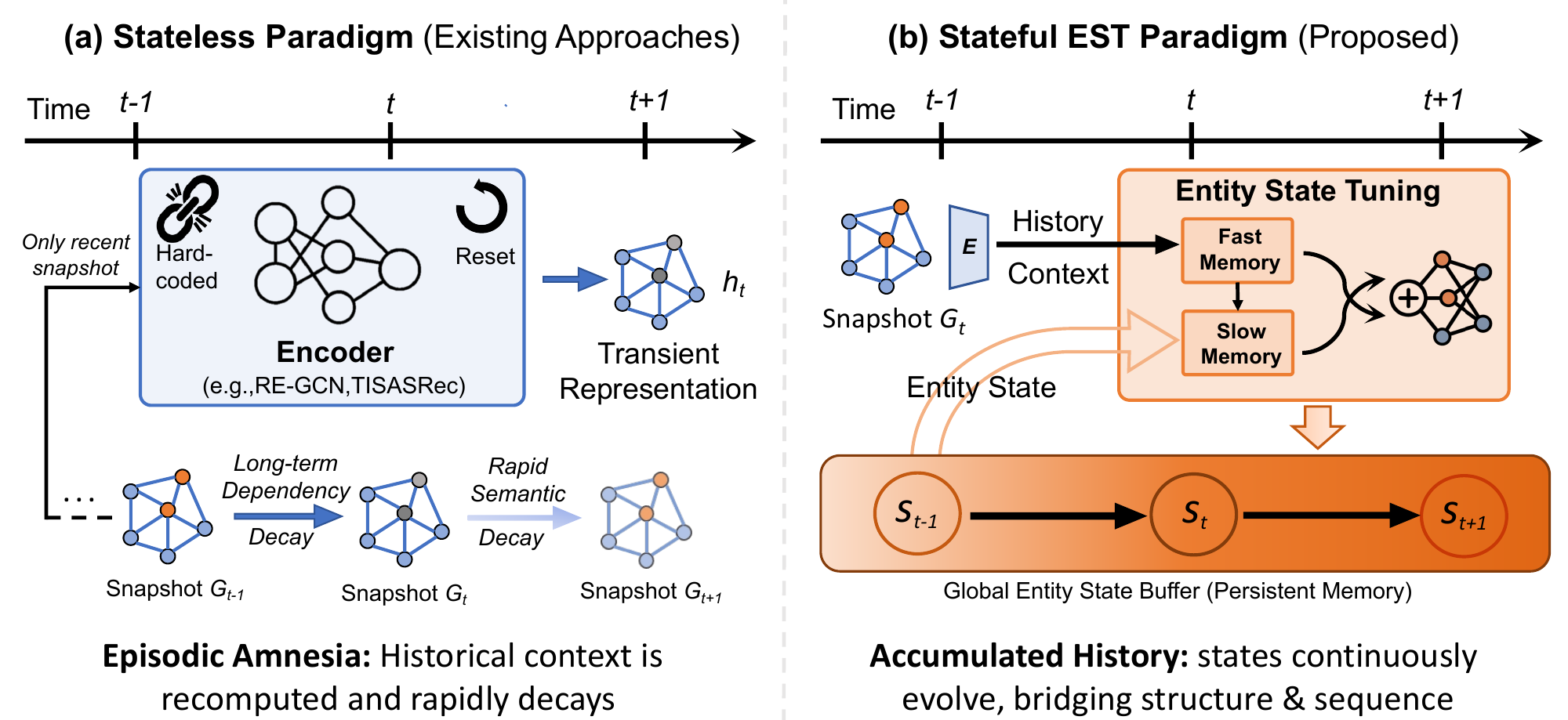}
    \caption{From Stateless Snapshot Encoding to Stateful Entity Reasoning in Temporal Knowledge Graphs.}
    \label{fig:est_paradigm}
\end{figure*}

Existing TKG forecasting methods typically follow a divide-and-conquer paradigm, focusing either on structural dependency modeling, exemplified by RE-GCN~\cite{li2021temporal}, or on sequential linearization, as in TiSASRec~\cite{li2020time}. 
Despite their architectural diversity, most of these methods still operate under a fundamentally stateless paradigm. 
As shown in Figure~\ref{fig:est_paradigm}(a), entity representations are reconstructed independently at each snapshot, preventing historical evidence from being persistently accumulated over time. 
As temporal reasoning proceeds, long-range dependencies are therefore progressively weakened, leading to what we term \textit{episodic amnesia}. 
By contrast, Figure~\ref{fig:est_paradigm}(b) highlights the necessity of modeling entities as persistent state carriers to naturally bridge structural and temporal dynamics.

Motivated by this observation, we propose an encoder-agnostic framework that redefines TKG reasoning through Entity State Tuning (EST). We contend that the key to superior performance lies not in the choice of a specific backbone, but in effectively bridging structural and sequential flows through persistent states. Specifically, we model entities as Dynamic State Containers that persist and evolve. Our framework integrates a modular structural encoder with a pluggable sequential encoder (ranging from classic RNNs to modern Mamba)~\cite{sherstinsky2020fundamentals,gu2024mamba}. Within this architecture, the EST mechanism acts as a semantic bridge: it continuously fuses topological context with temporal dynamics and writes the updates back into a global buffer. This ensures that entities evolve as stateful agents, preserving and accumulating historical knowledge along the timeline. Complementary to state tuning, we introduce Counterfactual Consistency Learning to correct observational visibility bias. Counterfactual constraints enable the model to isolate intrinsic temporal dependencies from spurious observational correlations.
Our main contributions are summarized as follows:
\begin{itemize}[noitemsep,nolistsep]

\item We identify the stateless bottleneck in existing TKG methods and pioneer a state-centric paradigm. This approach unifies structural and sequential dynamics by modeling entities as continuously evolving agents, preventing semantic decay over time.

\item We propose Entity State Tuning, a universal framework that empowers diverse backbones with persistent memory. EST effectively harmonizes topological context with temporal evolution, boosting performance without altering the core encoder architecture.

\item We introduce Counterfactual Consistency Learning to mitigate visibility bias. This objective steers the model away from spurious correlations, ensuring that the entity state evolution is grounded in temporal dependencies.

\item Extensive experiments demonstrate EST's state-of-the-art performance, empirically validating the framework's universality and the critical necessity of state persistence for long-horizon extrapolation.

\end{itemize}

\section{Related Works}
\label{sec:related_work}

\paragraph{Temporal Knowledge Graph Forecasting.} Existing TKG forecasting literature predominantly navigates the dichotomy between \textit{sequential evolution} and \textit{structural reasoning}~\cite{cai2022temporal,wang2023survey}. Sequence-centric approaches, exemplified by RE-NET~\cite{jin2020recurrent} and CyGNet~\cite{zhu2021learning}, linearize historical neighborhoods to capture repetitive temporal patterns via recurrent or copy-generation mechanisms. Conversely, structure-centric methods like RE-GCN~\cite{li2021temporal} and CEN~\cite{li2022complex} prioritize topological dependencies within snapshots, employing evolving GCNs to propagate information across dynamic neighbors. More recently, the field has expanded into generative and cognitive paradigms: DiffuTKG~\cite{cai2024predicting} leverages diffusion processes to model future stochasticity, while emerging works in 2025, such as CognTKE~\cite{chen2025cogntke} and TRCL~\cite{liu2025temporal}, incorporate cognitive dual-process theories and recursive contrastive learning to enhance extrapolation. However, a fundamental bottleneck persists: entities are treated as stateless ``tabula rasa,'' relying on bounded retrieval. EST transcends this by maintaining global persistent states, ensuring forecasting is driven by continuous evolutionary inertia rather than episodic context.

\paragraph{Stateful Reasoning in Dynamic Graphs.}
Maintaining persistent node states has been extensively explored in continuous-time dynamic graph learning. Models like JODIE~\cite{kumar2019predicting} and TGN~\cite{rossi2020temporal} utilize memory modules to store historical interaction patterns. Recent advances like DyMemR~\cite{zhang2024temporal} further extend this by employing dynamic memory pools to capture repetitive historical facts. Distinct from these mechanisms, which often serve as passive storage for raw events or homogeneous interactions, EST introduces a specialized framework for stateful TKG forecasting. Unlike standard architectures that decouple memory updates from topological aggregation, EST adopts a \textit{state-first} paradigm: it actively injects global state priors before structural encoding, creating a closed-loop evolution that mitigates semantic decay in long-term reasoning.

\section{Problem Formulation}
\label{sec.3}

Let $\mathcal{E}$ and $\mathcal{R}$ denote the sets of entities and relations, respectively. A Temporal Knowledge Graph (TKG) is formalized as a sequence of snapshots $\mathcal{G} = \{\mathcal{G}_1, \mathcal{G}_2, \dots, \mathcal{G}_T\}$, where each snapshot $\mathcal{G}_t = \{(s, r, o, t) \mid s, o \in \mathcal{E}, r \in \mathcal{R}\}$ comprises quadruples representing facts valid at timestamp $t$.

Given a future-time query $q = (s, r, ?, t)$ with $t > T_{\mathrm{obs}}$, the task of \emph{TKG extrapolation} aims to predict the missing object entity by leveraging the historical interactions of the subject $s$. The historical context of $s$ prior to time $t$ is defined as $\mathcal{H}_s(t) = \{ (o_k, r_k, \tau_k) \mid (s, r_k, o_k, \tau_k) \in \mathcal{G}_{\tau_k},\ \tau_k < t \}$, characterizing the temporal evolution of relational patterns associated with the subject.

In contrast to the prevailing \emph{stateless} paradigm which relies solely on discrete event retrieval, we postulate that each entity $e \in \mathcal{E}$ resides on a \emph{continuously evolving state manifold} within a latent semantic space. Let $\mathbf{S}_t \in \mathbb{R}^{|\mathcal{E}|\times d}$ denote the global entity state matrix at time $t$. Our key premise is that future prediction is conditional not only on the event history $\mathcal{H}_s(t)$ but critically on the entity's current evolutionary momentum, encapsulated by $\mathbf{S}_{t-1}$. Accordingly, the learning objective reduces to maximizing the log-likelihood:
\begin{equation}
\max_{\Theta} \sum_{(s,r,o,t)\in\mathcal{D}}
\log P_{\Theta}\!\left(o \mid s, r, \mathbf{S}_{t-1}, \mathcal{H}_s(t)\right).
\end{equation}
This formulation unifies historical structural evidence with evolving latent states, enabling the model to retain long-term inertia beyond the immediate context window.

\section{Methodology}
\label{sec.4}

 EST is founded on the principle of \emph{State Persistence}, comprising three tightly coupled components:
(1) a Topology-Aware State Perceiver that harmonizes local structural dependencies with global state priors;
(2) a Unified Temporal Context Module that synthesizes structural and temporal signals into a continuous sequence; and
(3) a De-confounded State Evolution mechanism that closes the loop by updating global states via a dual-system memory derived from temporal guarantees.

\subsection{Topology-Aware State Perceiver}

To bridge discrete structural events with continuous entity evolution, we adopt a \textit{State-First} encoding strategy. Unlike stateless GNNs, we posit that the interpretation of any topological dependency must be explicitly modulated by the participating entity's current state.

\paragraph{State-Augmented Entity Representation.}
For a neighboring entity $o_k$ involved in the $k$-th interaction, we synthesize a time-aligned representation $\tilde{\mathbf{E}}_{o_k}$ by fusing its invariant static embedding $\mathbf{E}_{o_k}$ with its dynamic state $\mathbf{S}_{o_k}$ retrieved from the global buffer. An adaptive gating mechanism mitigates semantic drift by dynamically balancing intrinsic identity against evolving context:
\begin{align}
\label{eq:2-3}
    \mathbf{g}_k &= \sigma\left( \mathbf{W}_{g} [\mathbf{E}_{o_k} \,\|\, \mathbf{S}_{o_k}] \right), \\
    \tilde{\mathbf{E}}_{o_k} &= \mathbf{g}_k \odot \mathbf{E}_{o_k} + (1 - \mathbf{g}_k) \odot \mathbf{S}_{o_k},
\end{align}
where $\sigma(\cdot)$ is the sigmoid function and $\mathbf{W}_g$ is a learnable projection.

\paragraph{Structural Context Encoding.}
We then model the local interaction driven by relation $r_k$. The state-augmented feature $\tilde{\mathbf{E}}_{o_k}$ is fed into a structural encoder $\Phi_{\text{struct}}$ alongside the relation embedding $\mathbf{E}_{r_k}$ to derive the event representation:
\begin{equation}
    \mathbf{X}_k = \Phi_{\text{struct}}\left(\tilde{\mathbf{E}}_{o_k}, \mathbf{E}_{r_k}\right).
\end{equation}
Here, $\Phi_{\text{struct}}$ serves as a generalized topological projector. In our implementation, we instantiate $\Phi_{\text{struct}}$ flexibly to accommodate varying graph sparsities, ensuring the interaction is grounded in both local topology and historical state inertia.

\subsection{Unified Temporal Context Encoder}
\label{sec:4.2}

Having derived the state-enhanced structural representations $\{\mathbf{X}_k\}$, we proceed to model the continuous-time dynamics of the subject's history. This module synthesizes structural and temporal contexts into a unified sequential manifold.

\paragraph{Temporal Calibration.}
To account for the irregularity of event intervals, we introduce a time-difference projection operator $\varphi(\Delta t)$. Let $\Delta \tau_k = t - \tau_k$ denote the elapsed time since the $k$-th interaction. The calibrated input $\mathbf{u}_k$ is defined as:
\begin{equation}
    \mathbf{u}_k = \mathrm{MLP}\!\left(
    \left[ \mathbf{X}_k \,\|\, \mathbf{E}_{r_k} \,\|\, \varphi(\Delta \tau_k) \right]
    \right).
\end{equation}
By explicitly re-injecting $\mathbf{E}_{r_k}$, we preserve relation-specific semantics that might be diluted during structural aggregation.

\paragraph{Pluggable Sequence Backbone.}
We formulate EST as an encoder-agnostic framework that supports diverse sequential architectures within the same state-tuning pipeline.
Given the stream $\mathbf{U} = [\mathbf{u}_1, \dots, \mathbf{u}_L]$, the sequence backbone maps each causal prefix to a contextualized representation:
\begin{equation}
    \mathbf{h}_k = \mathcal{T}\big(\mathbf{u}_{\le k}; \theta_{\text{seq}}\big), \quad
    \mathbf{y}_k = \mathcal{P}(\mathbf{h}_k),
\end{equation}
where $\mathbf{u}_{\le k} = [\mathbf{u}_1, \dots, \mathbf{u}_k]$ denotes the causally visible prefix up to step $k$, $\mathbf{h}_k$ is the contextualized representation produced by the chosen backbone at position $k$, and $\mathcal{P}(\cdot)$ is a projection head.
For notational uniformity, we use the same form for all backbone instantiations, while their concrete formulations (e.g., RNN, LSTM, Transformer, and Mamba) are provided in \textbf{Appendix~\ref{app:backbones}}.
We validate this flexibility empirically in Section~\ref{sec:5.2}.

\paragraph{Query-Specific Attention.}
To extract evidence relevant to the specific query $(s, r, ?, t)$, we apply a query-aware attention mechanism over the encoder output $\mathbf{Y} = [\mathbf{y}_1, \ldots, \mathbf{y}_L]$. The attention weight $\alpha_k$ is computed as:
\begin{equation}
\alpha_k \propto \exp\!\left(
\mathbf{v}^\top \tanh\!\left(
\mathbf{W}_{\text{att}}
\left[
\mathbf{y}_k;\,
\mathbf{e}_r;\,
\varphi(\Delta t)
\right]
\right)
\right).
\end{equation}
The final dynamic context $\mathbf{c}_s = \sum_{k=1}^{L} \alpha_k \mathbf{y}_k$ aggregates historical signals weighted by their relevance to the current prediction target.

\subsection{De-confounded State Evolution}
The core innovation of EST lies in its closed-loop state evolution. To resolve the \emph{Plasticity-Stability Dilemma} in non-stationary environments,
we draw inspiration from the Complementary Learning Systems theory in neuroscience~\cite{benna2016computational} to design a dual-track update scheme~\cite{kumaran2016learning, kirkpatrick2017overcoming}.

\paragraph{Fast System (Working Memory).}
We employ a high-sensitivity buffer $\mathbf{S}_{\text{fast}}$ to capture instantaneous semantic perturbations induced by the current context $\mathbf{c}_s$. This transient state is updated unconditionally via an exponential moving average:
\begin{equation}
\mathbf{S}_{\text{fast}}^{(t)} \leftarrow (1 - \lambda)\,\mathbf{S}_{\text{fast}}^{(t-1)} + \lambda\,\mathbf{c}_s,
\end{equation}
where $\lambda$ controls responsiveness. This process mimics the rapid reactivity of working memory to recent stimuli.

\paragraph{Slow System (Consolidated Memory).}
The global persistent state $\mathbf{S}_{\text{slow}}$ evolves only when significant structural shifts occur. We define the divergence between the fast and slow states as a measure of \emph{informational surprise}: $\delta_s = \| \mathbf{S}_{\text{fast}}^{(t)} - \mathbf{S}_{\text{slow}}^{(t-1)} \|_2$.
To suppress stochastic noise, we introduce an energy-barrier gating mechanism $g_s$:
\begin{equation}
\begin{aligned}
g_s &= \sigma\!\left( \kappa \cdot (\delta_s - \gamma) \right), \\
\mathbf{S}_{\text{slow}}^{(t)} &\leftarrow
\mathbf{S}_{\text{slow}}^{(t-1)} +
g_s \cdot
\left(
\mathbf{S}_{\text{fast}}^{(t)} -
\mathbf{S}_{\text{slow}}^{(t-1)}
\right),
\end{aligned}
\end{equation}
where $\gamma$ is the cognitive threshold. This ensures the entity manifold is reconfigured only when accumulated evidence surpasses the energy barrier ($\delta_s > \gamma$), distilling robust knowledge while filtering transient noise.

\paragraph{Counterfactual Consistency Learning.}
To prevent the model from capturing spurious correlations (e.g., merely memorizing high-frequency historical entities),
we employ a \textit{Counterfactual Negative Sampling} strategy to construct the negative set $\mathcal{E}^-$. Specifically, for a valid fact $(s, r, o, t)$, we sample negative entities $o'$ that are historically plausible (e.g., have interacted with $s$ in the past) but are factually incorrect at the current timestamp $t$.
The optimization is to minimize the following negative log-likelihood loss:
\begin{equation}
\begin{split}
\mathcal{L}_{\text{ccl}} &= - \operatorname*{\mathbb{E}}_{(s,r,o,t)\sim\mathcal{D}} \Big[ \phi(\mathbf{c}_s, r, o) - \\
&\quad \log \sum_{o' \in \{o\} \cup \mathcal{E}^-} \exp(\phi(\mathbf{c}_s, r, o')) \Big],
\end{split}
\end{equation}
where $\phi(\cdot)$ denotes the scoring function (variants detailed in \textbf{Appendix~\ref{app:scoring_function}}). This normalization over $\{o\} \cup \mathcal{E}^-$ compels the model to distinguish intrinsic temporal dependencies from historical priors.

\begin{table*}[t]
\centering
\small
\setlength{\tabcolsep}{3.6pt}
\renewcommand{\arraystretch}{1.12}
\caption{Overall performance on four benchmark datasets. Best results are in \textbf{bold}, second best are \underline{underlined}.}
\label{tab:main_result}
\resizebox{\textwidth}{!}{%
\begin{tabular}{@{}lcccccccccccccccc@{}}
\toprule
\multirow{2}{*}{\textbf{Model}} 
& \multicolumn{4}{c}{\textbf{ICEWS14}} 
& \multicolumn{4}{c}{\textbf{ICEWS18}} 
& \multicolumn{4}{c}{\textbf{ICEWS05-15}} 
& \multicolumn{4}{c}{\textbf{GDELT}} \\
\cmidrule(lr){2-5}\cmidrule(lr){6-9}\cmidrule(lr){10-13}\cmidrule(l){14-17}
& \textbf{MRR} & \textbf{Hits@1} & \textbf{Hits@3} & \textbf{Hits@10}
& \textbf{MRR} & \textbf{Hits@1} & \textbf{Hits@3} & \textbf{Hits@10}
& \textbf{MRR} & \textbf{Hits@1} & \textbf{Hits@3} & \textbf{Hits@10}
& \textbf{MRR} & \textbf{Hits@1} & \textbf{Hits@3} & \textbf{Hits@10} \\
\midrule
RE-NET (2020)       & 0.369 & 0.268 & 0.395 & 0.548 & 0.288 & 0.191 & 0.324 & 0.475 & 0.433 & 0.334 & 0.478 & 0.631 & 0.196 & 0.124 & 0.210 & 0.340 \\
CyGNet (2020)       & 0.351 & 0.257 & 0.390 & 0.536 & 0.249 & 0.159 & 0.283 & 0.426 & 0.368 & 0.266 & 0.416 & 0.562 & 0.185 & 0.115 & 0.196 & 0.320 \\
xERTE (2021)        & 0.400 & 0.321 & 0.446 & 0.562 & 0.293 & 0.210 & 0.335 & 0.465 & 0.466 & 0.378 & 0.523 & 0.639 & 0.181 & 0.123 & 0.201 & 0.303 \\
TITer (2021)        & 0.409 & 0.323 & 0.455 & 0.571 & 0.300 & 0.221 & 0.335 & 0.448 & 0.477 & 0.380 & 0.529 & 0.658 & 0.155 & 0.110 & 0.156 & 0.243 \\
RE-GCN (2021)       & 0.404 & 0.307 & 0.450 & 0.592 & 0.306 & 0.210 & 0.343 & 0.488 & 0.480 & 0.373 & 0.539 & 0.683 & 0.196 & 0.124 & 0.209 & 0.337 \\
CEN (2022)          & 0.422 & 0.321 & 0.475 & 0.613 & 0.315 & 0.217 & 0.354 & 0.506 & 0.468 & 0.364 & 0.524 & 0.670 & 0.204 & 0.130 & 0.218 & 0.350 \\
TiRGN (2022)        & 0.440 & 0.338 & 0.490 & 0.638 & 0.337 & 0.232 & 0.380 & 0.542 & 0.500 & 0.393 & 0.561 & 0.707 & 0.217 & 0.136 & 0.233 & 0.376 \\
HisMatch (2022)     & 0.464 & 0.359 & 0.516 & 0.668 & 0.340 & 0.239 & 0.379 & 0.539 & 0.528 & 0.420 & 0.591 & 0.733 & 0.220 & 0.145 & 0.238 & 0.366 \\
RETIA (2023)        & 0.428 & 0.323 & 0.478 & 0.628 & 0.324 & 0.222 & 0.365 & 0.529 & 0.473 & 0.366 & 0.529 & 0.678 & 0.201 & 0.128 & 0.215 & 0.345 \\
CENET (2023)        & 0.390 & 0.296 & 0.432 & 0.575 & 0.279 & 0.182 & 0.316 & 0.470 & 0.420 & 0.322 & 0.469 & 0.604 & 0.202 & 0.127 & 0.217 & 0.349 \\
PPT (2023)          & 0.384 & 0.289 & 0.425 & 0.570 & 0.266 & 0.169 & 0.306 & 0.454 & 0.388 & 0.286 & 0.434 & 0.586 & --    & --    & --    & --    \\
LogCL (2023)        & \underline{0.489} & \underline{0.378} & \textbf{0.547} & \textbf{0.703} & 0.357 & 0.245 & 0.403 & 0.577 & 0.570 & 0.461 & \underline{0.637} & \textbf{0.779} & 0.238 & 0.146 & 0.256 & 0.423 \\
Re-Temp (2023)      & 0.480 & 0.373 & 0.536 & \underline{0.689} & 0.358 & 0.250 & 0.404 & 0.573 & 0.563 & 0.455 & 0.628 & \underline{0.772} & 0.250 & 0.157 & 0.271 & 0.442 \\
DiffuTKG (2024)     & --    & --    & --    & --    & 0.367 & 0.257 & --    & 0.578 & 0.527 & 0.404 & --    & 0.760 & 0.251 & 0.162 & --    & 0.423 \\
CognTKE (2025)      & 0.461 & 0.365 & 0.511 & 0.645 & 0.352 & 0.252 & 0.399 & 0.547 & 0.531 & 0.426 & 0.594 & 0.727 & --    & --    & --    & --    \\
DSEP (2025)         & 0.449 & 0.344 & 0.503 & 0.649 & 0.338 & 0.232 & 0.382 & 0.545 & 0.507 & 0.398 & 0.568 & 0.713 & 0.220 & 0.139 & 0.237 & 0.381 \\
\midrule

\rowcolor{blue!6}
\textbf{EST-RNN}         & 0.501 & 0.447 & 0.524 & 0.618 & 0.458 & 0.398 & \underline{0.480} & 0.575 & \underline{0.611} & \underline{0.560} & 0.631 & 0.712 & \underline{0.408} & 0.348 & \underline{0.426} & \underline{0.510} \\

\rowcolor{blue!6}
\textbf{EST-LSTM}        & 0.505 & 0.448 & 0.532 & 0.625 & \textbf{0.467} & \textbf{0.407} & \textbf{0.490} & \textbf{0.585} & 0.609 & 0.558 & 0.629 & 0.710 & \underline{0.408} & \underline{0.350} & 0.425 & \textbf{0.519} \\

\rowcolor{blue!6}
\textbf{EST-Transformer} & \textbf{0.513} & \textbf{0.455} & 0.536 & 0.629 & \underline{0.466} & \underline{0.405} & \textbf{0.490} & \underline{0.583} & \textbf{0.621} & \textbf{0.572} & \textbf{0.640} & 0.718 & \textbf{0.412} & \textbf{0.356} & \textbf{0.429} & \textbf{0.519} \\

\rowcolor{blue!6}
\textbf{EST-Mamba}       & \underline{0.511} & \underline{0.451} & \underline{0.538} & 0.633 & 0.449 & 0.383 & 0.475 & 0.576 & 0.599 & 0.544 & 0.624 & 0.709 & 0.390 & 0.329 & 0.408 & 0.504 \\
\bottomrule
\end{tabular}%
}
\vspace{-1mm}
\end{table*}

\section{Experiments}

We empirically validate EST by addressing the following research questions:
\begin{itemize}[leftmargin=*, itemsep=0pt, topsep=0pt]
    \item \textbf{RQ1:} How does EST perform compared to representative TKG models?
    \item \textbf{RQ2:} What are the contributions of EST’s key components, and how sensitive is the model to key hyperparameters?
    \item \textbf{RQ3:} How does the EST model the evolution of entity states over time, and how do structural and sequential signals interact within this process?
    \item \textbf{RQ4:} Can the learned entity states generalize across time beyond the training distribution?
\end{itemize}

\subsection{Experiment Settings}

\paragraph{Datasets.}
We evaluate EST on four standard benchmarks—\textbf{ICEWS14}, \textbf{ICEWS18}, \textbf{ICEWS05-15}, and \textbf{GDELT}—spanning diverse scales of political and global event dynamics. Detailed dataset statistics and implementation details are provided in \textbf{Appendix~\ref{app:datasets} and ~\ref{app:implementation_details}}.

\paragraph{Baselines.}
We compare against a broad set of TKG models, including RE-NET\cite{jin2020recurrent}, CyGNet\cite{zhu2021learning}, xERTE\cite{han2020explainable}, TIter\cite{sun2021timetraveler}, RE-GCN\cite{li2021temporal}, CEN\cite{li2022complex}, TiRGN\cite{li2022tirgn},
HiSMatch\cite{li2022hismatch}, RETIA\cite{liu2023retia}, CENET\cite{xu2023temporal}, PPT\cite{xu2023pre}, LogCL\cite{chen2024local}, Re-Temp\cite{wang2023re}, DiffuTKG\cite{cai2024predicting}, CognTKE\cite{chen2025cogntke}, and DSEP\cite{deng2025diachronic}.

\paragraph{Evaluation Metrics.}
We adopt the standard filtered MRR and Hits@N (H@N) for evaluation.

\begin{table*}[ht]
\centering
\footnotesize
\setlength{\tabcolsep}{5.5pt}
\renewcommand{\arraystretch}{1.0}
\caption{Cumulative wall-clock training time (minutes) at different training epochs on the ICEWS14 dataset.}
\begin{tabular}{l|ccccccccccccc}
\toprule
\textbf{Model} & \multicolumn{13}{c}{\textbf{Training Epochs}} \\
\cmidrule(lr){2-14}
& \textbf{Start} & \textbf{1} & \textbf{2} & \textbf{3} & \textbf{4} & \textbf{5} & \textbf{6} & \textbf{7} & \textbf{8} & \textbf{9} & \textbf{10} & \textbf{15} & \textbf{20} \\
\midrule
Ours w/ RNN         & 0.00 & 1.07 & 2.12 & 3.17 & 4.18 & 5.22 & 6.27 & 7.32 & 8.35 & 9.38 & 10.42 & 15.62 & 20.85 \\
Ours w/ LSTM        & 0.00 & 1.13 & 2.22 & 3.32 & 4.40 & 5.48 & 6.60 & 7.70 & 8.78 & 9.88 & 11.00 & 16.52 & 22.03 \\
Ours w/ Transformer & 0.00 & 0.95 & 1.90 & 2.82 & 3.75 & 4.68 & 5.62 & 6.55 & 7.48 & 8.42 & 9.35 & 13.92 & 18.58 \\
Ours w/ Mamba       & 0.00 & 1.03 & 2.02 & 3.01 & 4.00 & 5.00 & 5.98 & 6.97 & 7.98 & 8.95 & 9.95 & 14.90 & 19.81 \\
\bottomrule
\end{tabular}
\label{tab:train_time}
\end{table*}

\begin{figure*}[h]
  \centering
  \begin{subfigure}[t]{0.48\linewidth}
    \centering
    \includegraphics[width=\linewidth]{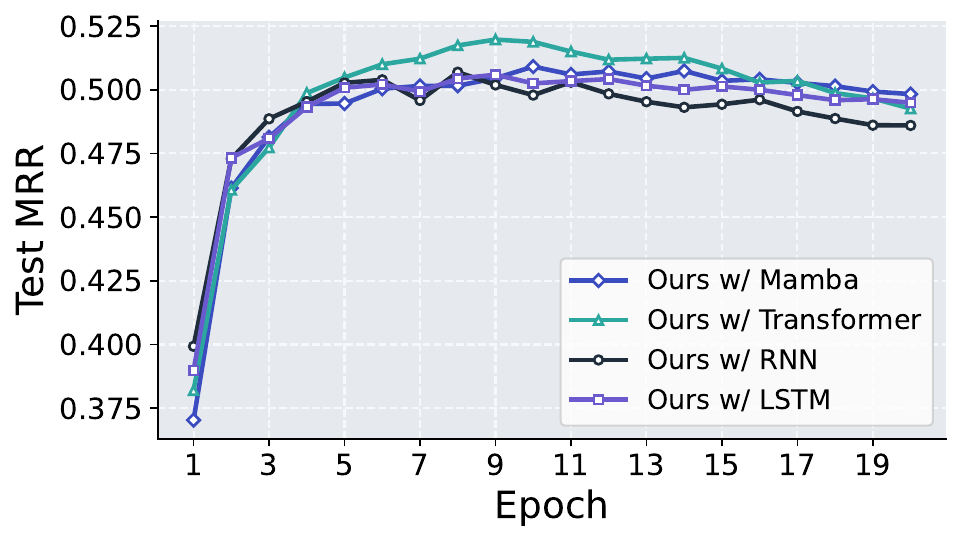}
    \caption{Test MRR convergence.}
    \label{fig:encoder_mrr_convergence}
  \end{subfigure}
  \hfill
  \begin{subfigure}[t]{0.47\linewidth}
    \centering
    \includegraphics[width=\linewidth]{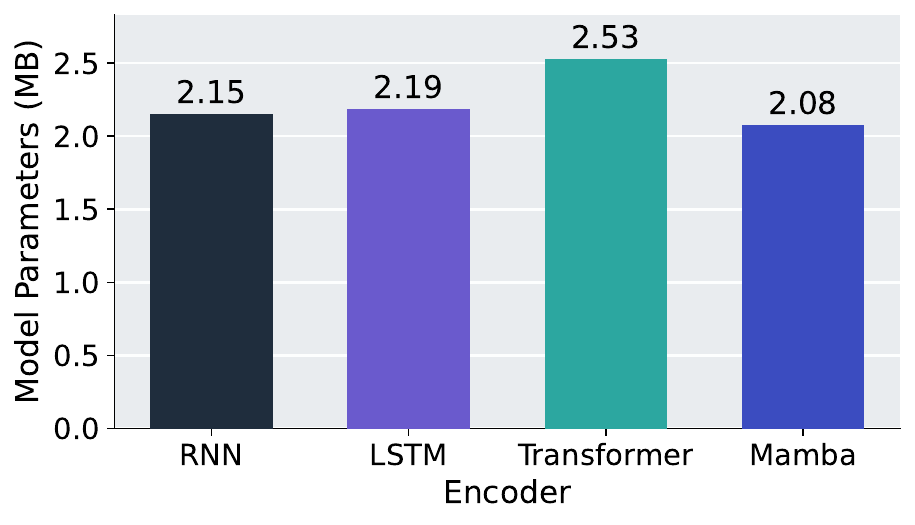}
    \caption{Model size comparison (MB).}
    \label{fig:encoder_params_mb}
  \end{subfigure}
  \caption{Efficiency comparison among different sequence encoders on the ICEWS14 dataset.}
  \label{fig:encoder_efficiency}
  \vspace{-4mm}
\end{figure*}

\begin{table}[t]
\centering
\small
\caption{Additional memory introduced by the two FP32 global state buffers in EST ($d{=}64$).}
\label{tab:buffer_memory}
\begin{tabular}{lrr}
\toprule
\textbf{Dataset} & \textbf{\#Entities} & \textbf{Buffer memory (FP32)} \\
\midrule
ICEWS14    & 7,128   & 3.48 MB \\
ICEWS18    & 10,094  & 4.93 MB \\
ICEWS05-15 & 23,033  & 11.25 MB \\
GDELT      & 7,691   & 3.76 MB \\
\midrule
Extrapolation & 1,000,000 & 488.28 MB ($\sim$0.49 GB) \\
\bottomrule
\end{tabular}
\vspace{-2mm}
\end{table}

\begin{table}[t]
\centering
\small
\caption{Measured runtime and peak-memory overhead of EST on ICEWS14, compared with a matched stateless ablation (\textbf{w/o State}).}
\label{tab:state_overhead}
\begin{tabular}{lccc}
\toprule
\textbf{Setting} & \textbf{s/epoch} & \textbf{ms/step} & \textbf{Peak MB} \\
\midrule
EST-Mamba & 60.77 & 105.09 & 173.80 \\
w/o State & 59.60 & 102.46 & 165.21 \\
$\Delta$ & +1.96\% & +2.57\% & +5.20\% \\
\midrule
EST-Transformer & 61.51 & 105.22 & 111.61 \\
w/o State & 61.04 & 104.96 & 101.12 \\
$\Delta$ & +0.77\% & +0.25\% & +10.37\% \\
\bottomrule
\end{tabular}
\vspace{-4mm}
\end{table}

\subsection{Overall Performance Comparison: RQ1}
\label{sec:5.2}

\paragraph{Main results.}
Table~\ref{tab:main_result} demonstrates that the proposed EST framework consistently establishes new state-of-the-art results across all benchmarks, with EST-Transformer achieving an MRR of 0.513 on ICEWS14 and a remarkable 0.412 on GDELT. Most notably, on the high-noise GDELT dataset, our model yields a relative improvement of over \textbf{60\%} compared to leading baselines (e.g., DiffuTKG, Re-Temp), strongly validating that our \textit{persistent state mechanism} effectively mitigates the ``episodic amnesia'' inherent in stateless methods by robustly filtering long-term noise. Crucially, despite utilizing a lightweight MLP for structural encoding ($\Phi_{\text{struct}}$), even the basic EST-RNN variant outperforms structurally complex models (e.g., TiRGN, DSEP). This underscores that the bottleneck of TKG forecasting lies in the reasoning paradigm rather than encoder complexity. The consistent superiority across diverse backbones (from RNN to Mamba) further confirms EST as a universal solution for temporal reasoning.

\subsubsection{Efficiency and Scalability}

EST is encoder-agnostic, abstracting temporal evolution as a generalized state transition function $\mathcal{T}$.
We instantiate $\mathcal{T}$ with four backbones; the corresponding formulations are provided in \textbf{Appendix~\ref{app:backbones}}.
As shown in Figure~\ref{fig:encoder_efficiency} and Table~\ref{tab:train_time}, all variants converge reliably, indicating that the gains primarily come from persistent state modeling rather than a particular sequence encoder.
Among them, EST-Transformer achieves the best overall accuracy, while EST-Mamba offers the most parameter-efficient alternative with competitive performance.
Accordingly, we use EST-Transformer as the default accuracy-oriented instantiation and EST-Mamba as a lightweight counterpart in subsequent analyses.

Beyond encoder-level efficiency, we further analyze the overhead introduced by persistent global states.
EST maintains two \emph{non-trainable} entity buffers, $\mathbf{S}_{\mathrm{fast}}$ and $\mathbf{S}_{\mathrm{slow}}$, each of size $|\mathcal{E}| \times d$.
The resulting memory overhead is therefore linear in the number of entities, i.e., $\mathcal{O}(|\mathcal{E}|d)$.
Under our default setting ($d{=}64$, FP32), the total buffer footprint is
$
2 \times |\mathcal{E}| \times 64 \times 4 = 512|\mathcal{E}| \text{ bytes},
$
which remains at MB scale on standard benchmarks (Table~\ref{tab:buffer_memory}).

Importantly, EST does not scan all entity states at each step.
State access is sparse: for each batch, we only read and update entities appearing in the current history window.
As a result, the runtime remains dominated by the sequential backbone and local structural encoding, while the global buffers introduce only lightweight indexing and in-place updates.
Accordingly, the overall sequence modeling complexity remains that of the chosen backbone, namely $\mathcal{O}(Ld^2)$ for RNN/LSTM, $\mathcal{O}(L^2d)$ for Transformer, and $\mathcal{O}(Ld)$ for Mamba.

To quantify the practical overhead of persistent states, we compare EST against a strictly matched stateless ablation (\textbf{w/o State}) on ICEWS14.
Table~\ref{tab:state_overhead} shows that the additional runtime is modest for both backbones: for Mamba, EST introduces only $+1.96\%$ s/epoch and $+2.57\%$ ms/step, while for Transformer the increase is $+0.77\%$ and $+0.25\%$, respectively.
The peak-memory increase is also limited ($+5.20\%$ for Mamba and $+10.37\%$ for Transformer).
Overall, EST introduces a linear memory overhead in $|\mathcal{E}|$ and only minor runtime cost in practice, making the global-buffer design practical on standard TKGs and viable at larger scales.





\begin{table*}[ht]
\centering
\small
\setlength{\aboverulesep}{0pt}
\setlength{\belowrulesep}{0pt}
\renewcommand{\arraystretch}{1.1}
\setlength{\tabcolsep}{5pt}

\caption{Ablation study of key components on the ICEWS14 and ICEWS18 datasets.}
\label{tab:ablation}

\begin{subtable}[t]{0.45\linewidth}
\centering
\footnotesize
\begin{tabular}{l|cccc}
\toprule
\textbf{Model} & \textbf{MRR} & \textbf{Hits@1} & \textbf{Hits@3} & \textbf{Hits@10} \\ \midrule
\rowcolor[HTML]{EFEFEF}
\multicolumn{5}{c}{\textbf{ICEWS14}} \\ \hline
EST-Mamba & \textbf{0.511} & \textbf{0.451} & \textbf{0.538} & \textbf{0.633} \\
w/o Context & 0.360 & 0.266 & 0.403 & 0.536 \\
w/o State & 0.452 & 0.384 & 0.481 & 0.587 \\ 
w/o CCL & 0.489 & 0.427 & 0.511 & 0.613 \\
\hline
\rowcolor[HTML]{EFEFEF}
\multicolumn{5}{c}{\textbf{ICEWS18}} \\ \hline EST-Mamba & \textbf{0.449} & \textbf{0.383} & \textbf{0.475} & \textbf{0.576} \\
w/o Context & 0.259 & 0.165 & 0.295 & 0.442 \\
w/o State & 0.381 & 0.308 & 0.412 & 0.521 \\
w/o CCL & 0.433 & 0.380 & 0.468 & 0.557 \\
\bottomrule
\end{tabular}
\end{subtable}
\hfill
\begin{subtable}[t]{0.5\linewidth}
\centering
\footnotesize
\begin{tabular}{l|cccc}
\toprule
\textbf{Model} & \textbf{MRR} & \textbf{Hits@1} & \textbf{Hits@3} & \textbf{Hits@10} \\ \midrule
\rowcolor[HTML]{EFEFEF}
\multicolumn{5}{c}{\textbf{ICEWS14}} \\ \hline
EST-Transformer & \textbf{0.513} & \textbf{0.455} & \textbf{0.536} & \textbf{0.629} \\
w/o Context & 0.360 & 0.266 & 0.403 & 0.536 \\
w/o State & 0.455 & 0.386 & 0.479 & 0.582 \\ 
w/o CCL & 0.502 & 0.445 & 0.531 & 0.615 \\
\hline 
\rowcolor[HTML]{EFEFEF}
\multicolumn{5}{c}{\textbf{ICEWS18}} \\ \hline EST-Transformer & \textbf{0.466} &\textbf{ 0.405} & \textbf{0.490}& \textbf{0.583} \\
w/o Context & 0.259 & 0.165 & 0.295 & 0.442 \\
w/o State & 0.397 & 0.326 & 0.424 & 0.529 \\
w/o CCL & 0.449 & 0.396 & 0.487 & 0.559 \\
\bottomrule
\end{tabular}
\end{subtable}
\end{table*}

\begin{figure*}[ht]
    \centering
    \small
    \setlength{\tabcolsep}{0pt}

    \begin{subfigure}[b]{0.3\textwidth}
        \centering
        \includegraphics[width=\textwidth]{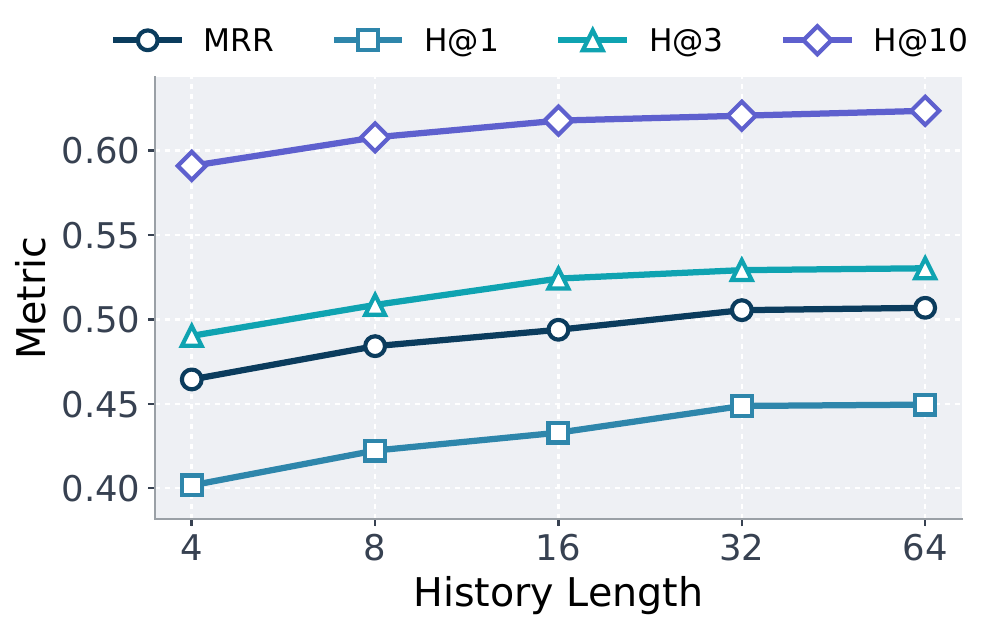}
        
    \end{subfigure}%
    \hfill
    \begin{subfigure}[b]{0.3\textwidth}
        \centering
        \includegraphics[width=\textwidth]{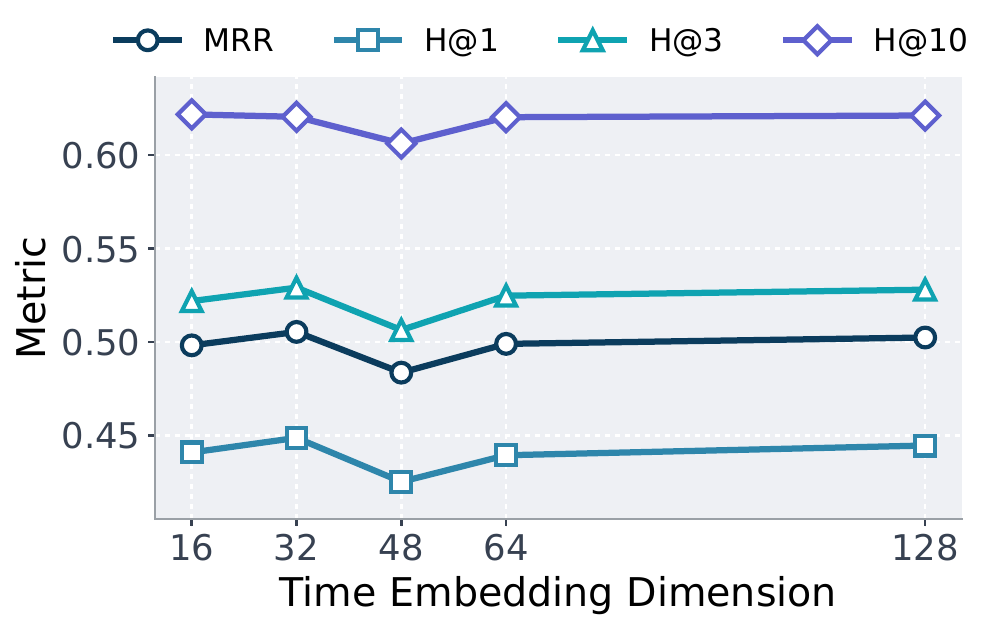}
        
    \end{subfigure}%
    \hfill
    \begin{subfigure}[b]{0.3\textwidth}
        \centering
        \includegraphics[width=\textwidth]{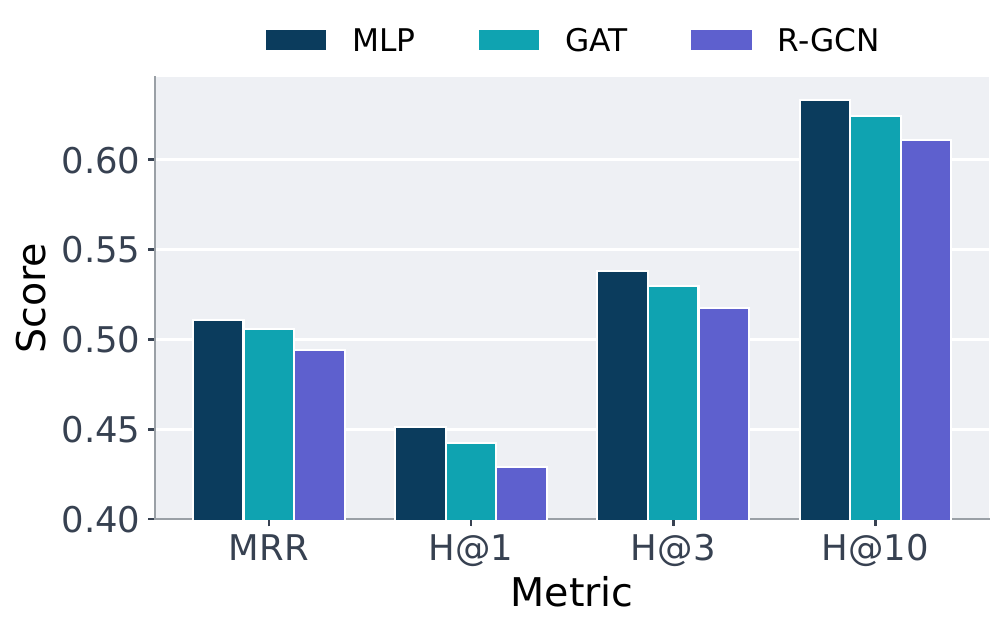}
        
    \end{subfigure}

    \vspace{0.2em}

    \begin{subfigure}[b]{0.3\textwidth}
        \centering
        \includegraphics[width=\textwidth]{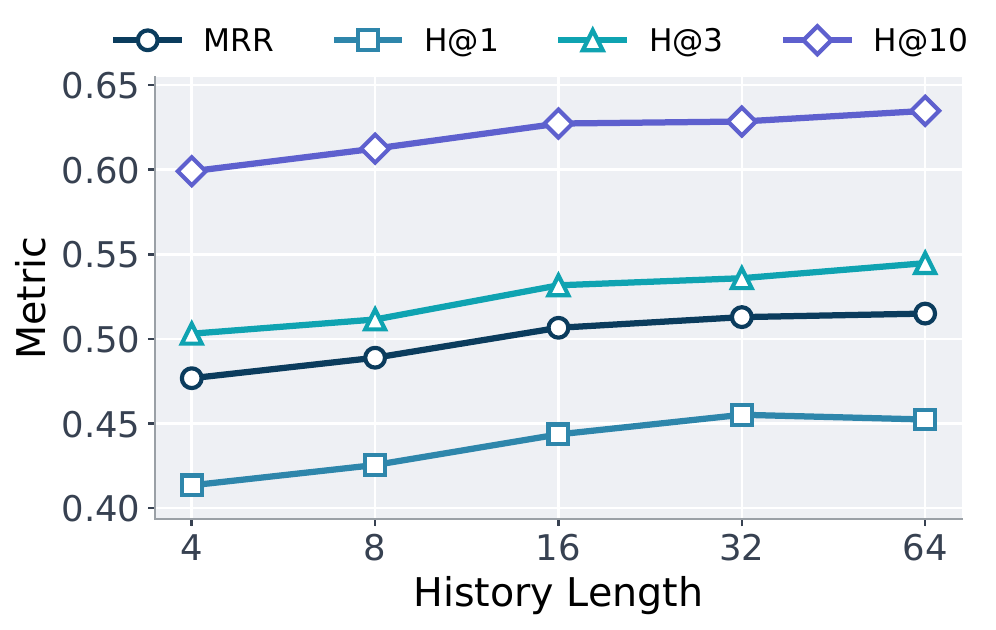}
        \caption{Impact of History Length}
    \end{subfigure}%
    \hfill
    \begin{subfigure}[b]{0.3\textwidth}
        \centering
        \includegraphics[width=\textwidth]{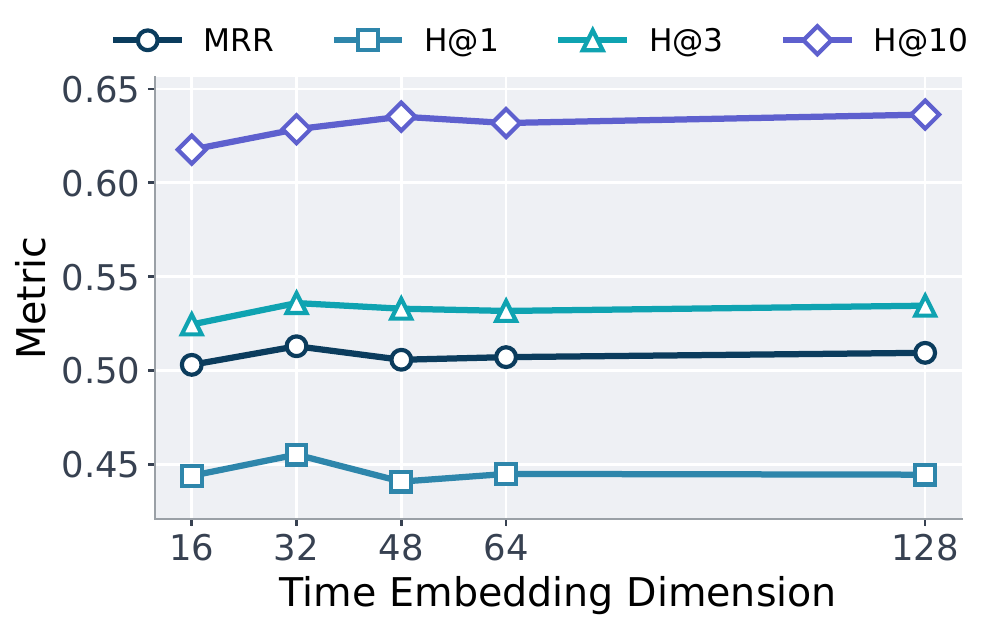}
        \caption{Impact of Time Embedding}
    \end{subfigure}%
    \hfill
    \begin{subfigure}[b]{0.3\textwidth}
        \centering
        \includegraphics[width=\textwidth]{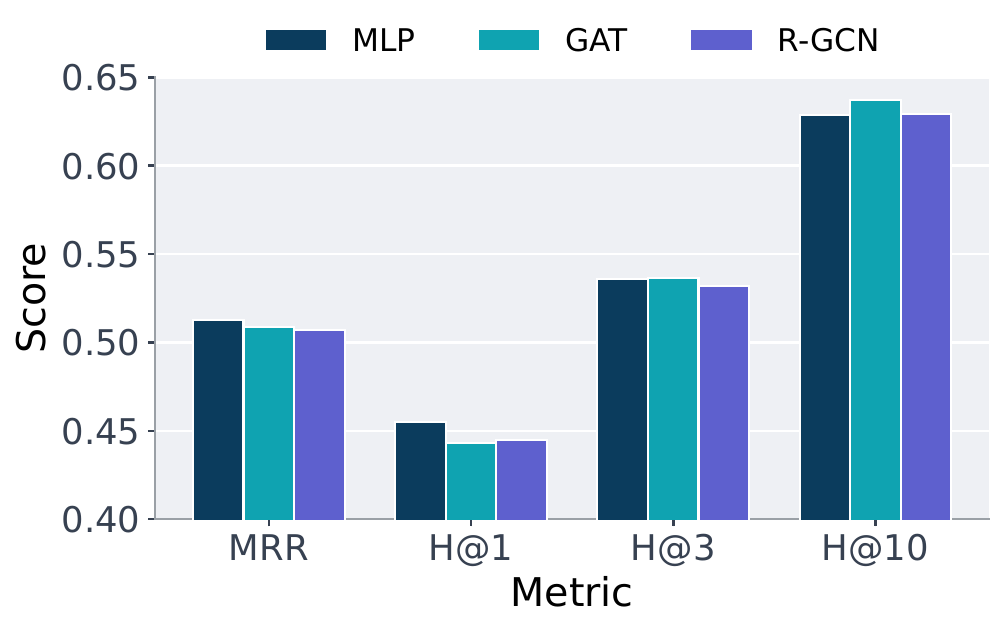}
        \caption{Impact of Structural Encoders}
    \end{subfigure}
    \caption{Hyperparameter analysis for EST-Mamba (top) and EST-Transformer (bottom).}
    \label{fig:hyperparam}
    \vspace{-2mm}
\end{figure*}

\subsection{Ablation Study: RQ2}

To verify the effectiveness of our EST, we develop three distinct model variants for ablation studies:
\begin{itemize}[leftmargin=*, itemsep=0pt, topsep=0pt]
    \item \textbf{w/o Context}: We remove both structural and sequential encoders, predicting from the query $(s,r, ?, t)$ alone.
    \item \textbf{w/o State}: We keep the context encoders but disable state retrieval and closed-loop state updates, using only static embeddings.
    \item \textbf{w/o CCL}: We remove the counterfactual consistency objective and train with the standard likelihood loss using conventional negative sampling.
\end{itemize}

\subsubsection{Key Modules Ablation Study}

Table~\ref{tab:ablation} dissects the impact of individual components across ICEWS14 and ICEWS18. We observe a clear hierarchical dependency in component efficacy. Removing the historical context (\textbf{w/o Context}) causes a precipitous performance drop (e.g., MRR declines from 0.511 to 0.360 on ICEWS14), confirming the foundational role of sequential evidence over static query matching. Crucially, decoupling the state persistence mechanism (\textbf{w/o State}) leads to a substantial degradation (average $\Delta$MRR $\approx$ -0.07) across all settings. This empirical gap directly validates our core hypothesis: without the persistent state to bridge temporal disjoints, the model suffers from rapid semantic decay. Furthermore, the removal of counterfactual consistency (\textbf{w/o CCL}) results in a consistent performance dip (up to 0.022 MRR), underscoring its necessity as a regularizer to filter spurious historical priors. 
These trends persist across Mamba and Transformer backbones, indicating EST’s backbone-agnostic design.

\begin{figure*}[ht]
\centering
\begin{minipage}[t]{0.49\textwidth}
\centering
\captionsetup{type=table}
\caption{Top-5 entities exhibiting the largest state displacement on the ICEWS14 dataset.}
\label{tab:top5_displacement_1}
\small
\setlength{\tabcolsep}{6pt}
\renewcommand{\arraystretch}{1.02}
\begin{adjustbox}{width=\linewidth}
\begin{tabular}{c >{\raggedright\arraybackslash}p{4.8cm} r r}
\toprule
\textbf{Rank} & \textbf{Entity} & \textbf{Gate $\mu$} & \textbf{Gate $\sigma$} \\
\midrule
1 & Men\_(Namibia)                  & 0.8875 & 0.1507 \\
2 & Political\_Parties\_(Indonesia) & 0.8742 & 0.1737 \\
3 & Police\_(Kosovo)                & 0.8865 & 0.1435 \\
4 & Criminal\_(Brazil)              & 0.8702 & 0.1800 \\
5 & Will\_Hodgman                   & 0.9126 & 0.1487 \\
\bottomrule
\end{tabular}
\end{adjustbox}
\end{minipage}
\hfill
\begin{minipage}[t]{0.49\textwidth}
\centering
\captionsetup{type=table}
\caption{Top-5 entities exhibiting the largest state displacement on the ICEWS18 dataset.}
\label{tab:top5_displacement_2}
\small
\setlength{\tabcolsep}{6pt}
\renewcommand{\arraystretch}{1.02}
\begin{adjustbox}{width=\linewidth}
\begin{tabular}{c >{\raggedright\arraybackslash}p{4.8cm} r r}
\toprule
\textbf{Rank} & \textbf{Entity} & \textbf{Gate $\mu$} & \textbf{Gate $\sigma$} \\
\midrule
1 & Police (Honduras)      & 0.8740 & 0.1653 \\
2 & Protester (Honduras)  & 0.8638 & 0.1767 \\
3 & Police (Malawi)       & 0.8698 & 0.1549 \\
4 & Protester (Tunisia)   & 0.8837 & 0.1573 \\
5 & Citizen (Malawi)      & 0.8721 & 0.1762 \\
\bottomrule
\end{tabular}
\end{adjustbox}
\end{minipage}

\vspace{3pt}

\captionsetup{type=figure} 

\begin{subfigure}[t]{0.45\textwidth}
    \centering
    \includegraphics[width=\linewidth]{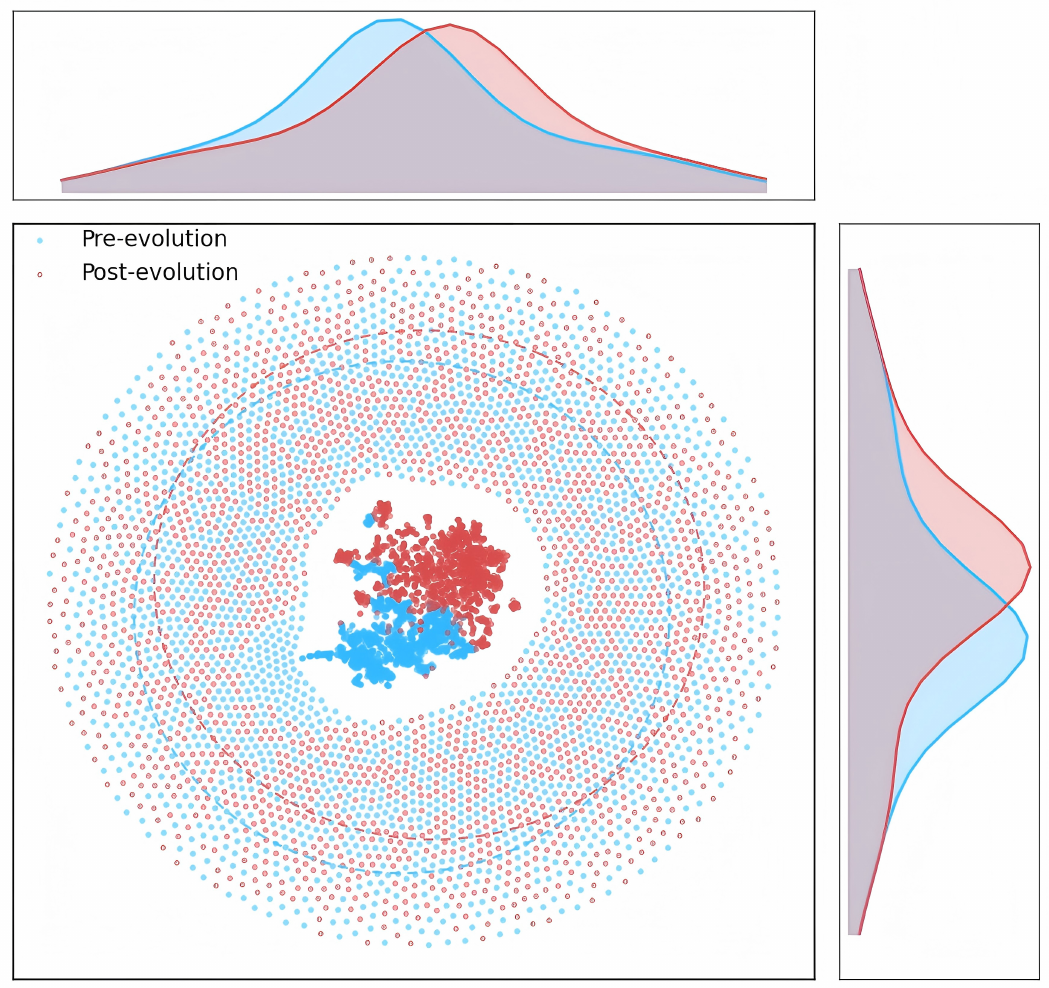}
    \label{fig:rq3_tsne_14}
\end{subfigure}
\hfill
\begin{subfigure}[t]{0.45\textwidth}
    \centering
    \includegraphics[width=\linewidth]{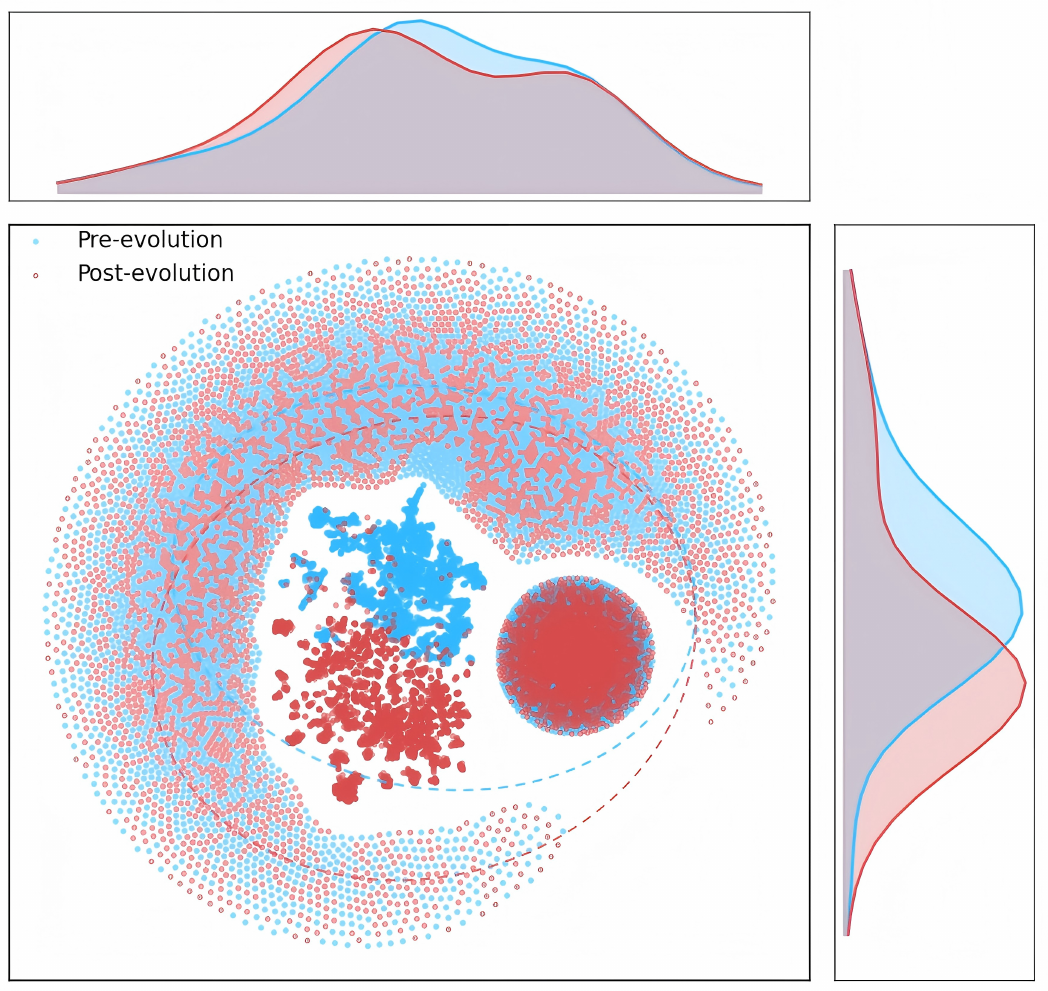}
    \label{fig:rq3_tsne_18}
\end{subfigure}
\vspace{-6pt}
\caption{t-SNE visualizations of entity states before and after evolution on ICEWS14 (left) and ICEWS18 (right).}
\label{fig:tsne_state}
\end{figure*}

\subsubsection{Sensitivity to Key Hyperparameters}

Figure~\ref{fig:hyperparam} studies the effects of history length $L$, time-embedding dimension, and structural encoders on ICEWS14.
Overall, both variants show consistent trends, indicating that EST is stable to these design choices.

\textbf{History length.} Increasing $L$ steadily improves MRR and Hits, with gains saturating around $L\!=\!32$; longer histories bring marginal benefits. This suggests EST can effectively exploit longer context, while remaining robust under short windows.

\textbf{Time embedding dimension.} Performance is largely insensitive to the time-embedding size. Both models peak around moderate dimensions (e.g., 32) and remain stable thereafter, with only minor fluctuations at intermediate sizes.

\textbf{Structural encoders.} Swapping the structural module among MLP, GAT, and R-GCN yields only small differences. A lightweight MLP is already competitive (often best on MRR), implying that the dominant improvements are driven by EST’s state persistence and temporal modeling rather than a heavily engineered structural encoder.

\subsection{Entity State Evolution Analysis: RQ3}

To visualize how entity states evolve over time, we track the evolution trajectories of a random 50\% subset of entities and project their persistent states with t-SNE~\cite{maaten2008visualizing}.
As shown in Figure~\ref{fig:tsne_state}, pre- and post-evolution states largely overlap for most entities, while a relatively small portion exhibits clear displacement and local cluster reconfiguration.
This pattern indicates that EST preserves a stable global geometry and selectively adjusts states when new evidence accumulates.

Motivated by the above, we quantify \emph{state displacement} as the $\ell_2$ change of an entity’s persistent state across updates and extract the top movers.
Tables~\ref{tab:top5_displacement_1}--\ref{tab:top5_displacement_2} list the top-5 entities with the largest displacement on ICEWS14/18, which correspond to highly eventful actors (e.g., \textit{Police/Protester/Citizen} in specific countries).
For these entities, we further report the statistics (Gate $\mu$/$\sigma$) of the \emph{state-augmentation gate} $\mathbf{g}_k$ in Eq.~\eqref{eq:2-3}, which controls how static identity embeddings and dynamic states are fused when constructing state-aware neighborhood representations.
The consistently high Gate $\mu$ with moderate variance suggests that, even for rapidly changing entities, EST keeps representations anchored to intrinsic identity while injecting state information in a controlled manner, mitigating semantic drift.
Overall, the t-SNE geometry and top-displacement analysis jointly illustrate a closed interaction loop: evolved states shape future structural encoding, which in turn drives subsequent state updates.

\begin{figure*}[t]
  \centering
  \begin{subfigure}[t]{0.48\linewidth}
    \centering
    \includegraphics[width=\linewidth]{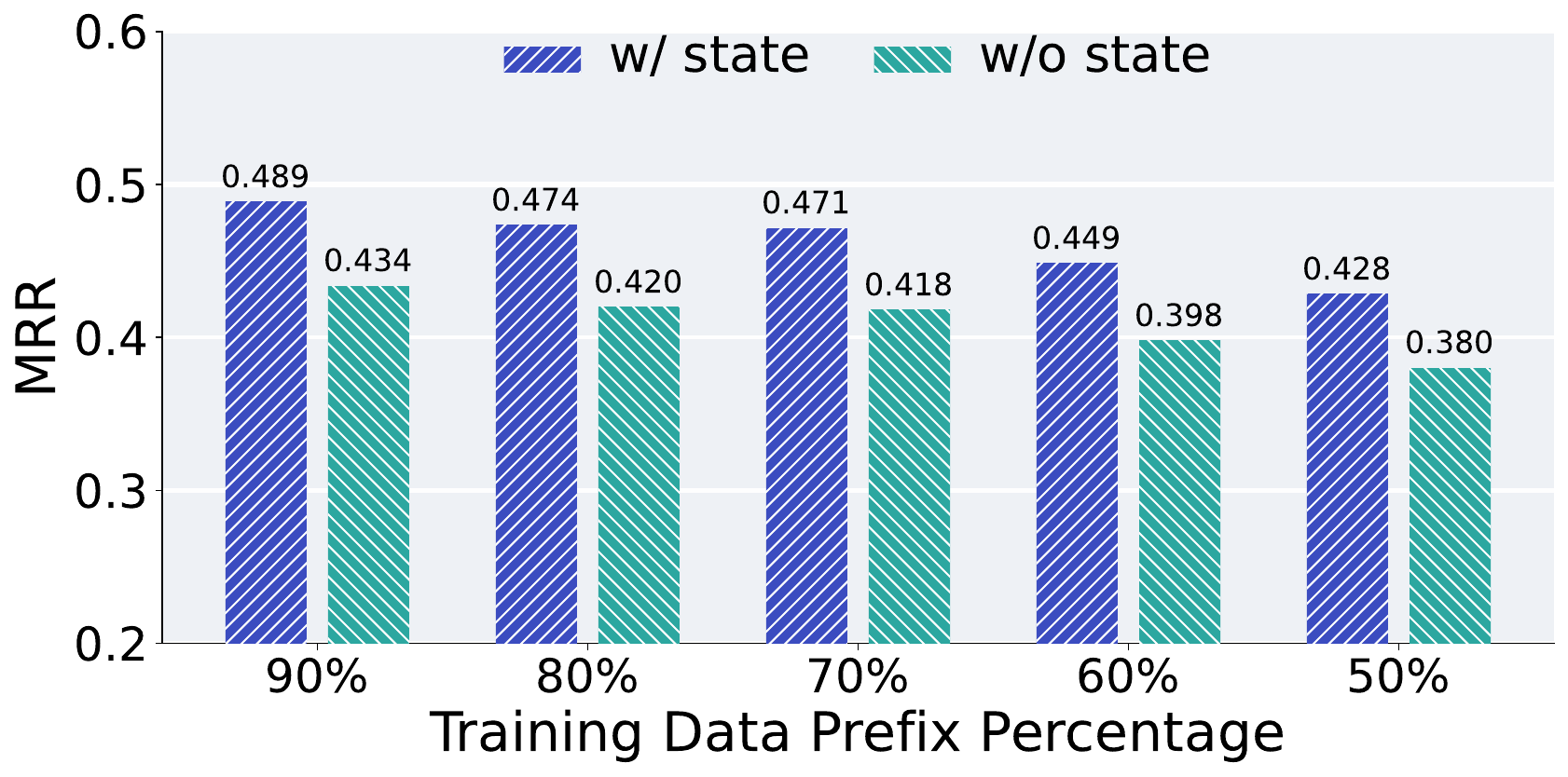}
    \caption{EST-Transformer on ICEWS14.}
    \label{fig:rq4_temporal_tr_icews14}
  \end{subfigure}
  \hfill
  \begin{subfigure}[t]{0.48\linewidth}
    \centering
    \includegraphics[width=\linewidth]{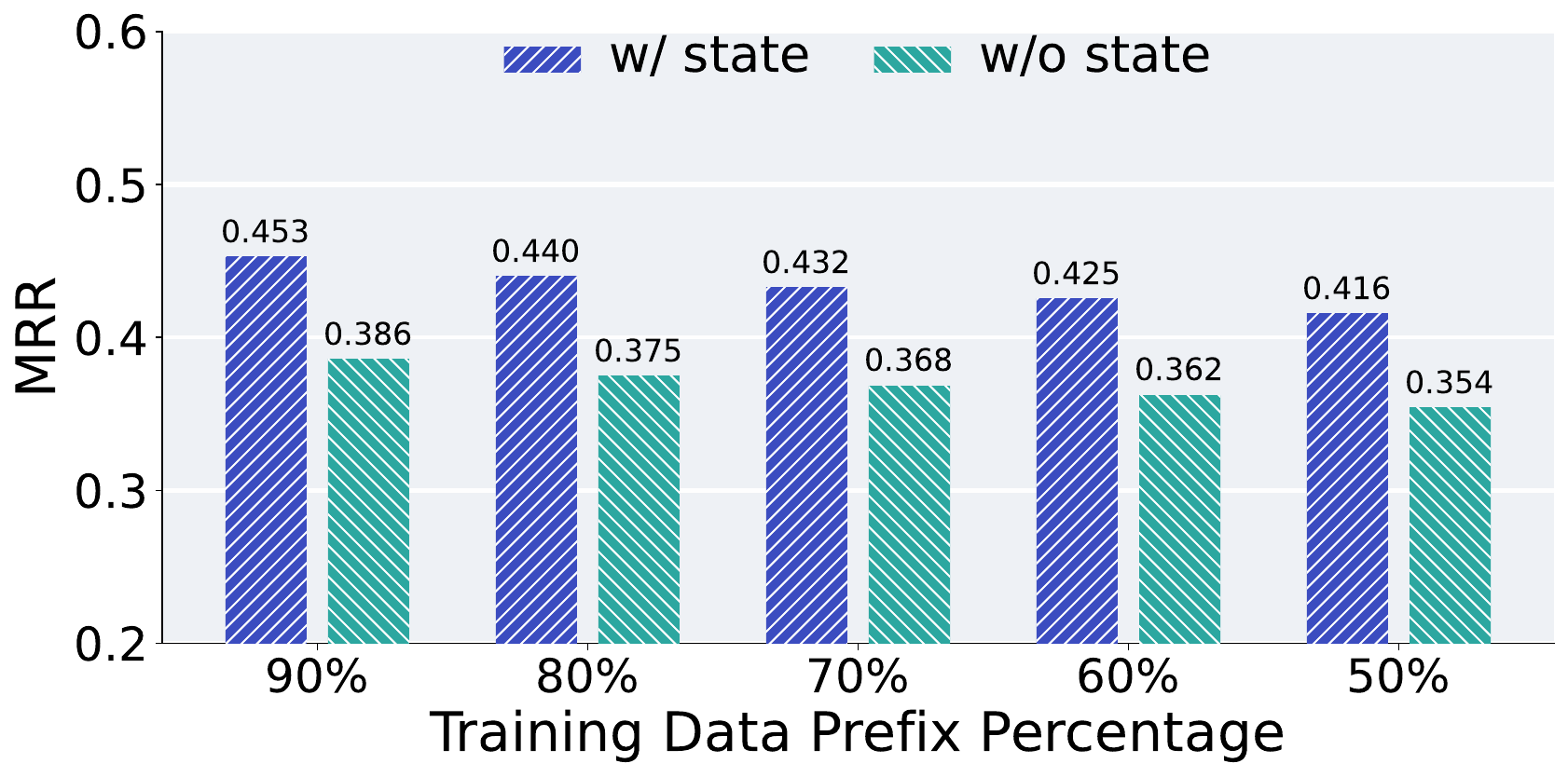}
    \caption{EST-Transformer on ICEWS18.}
    \label{fig:rq4_temporal_tr_icews18}
  \end{subfigure}

  \vspace{4pt}

  \begin{subfigure}[t]{0.48\linewidth}
    \centering
    \includegraphics[width=\linewidth]{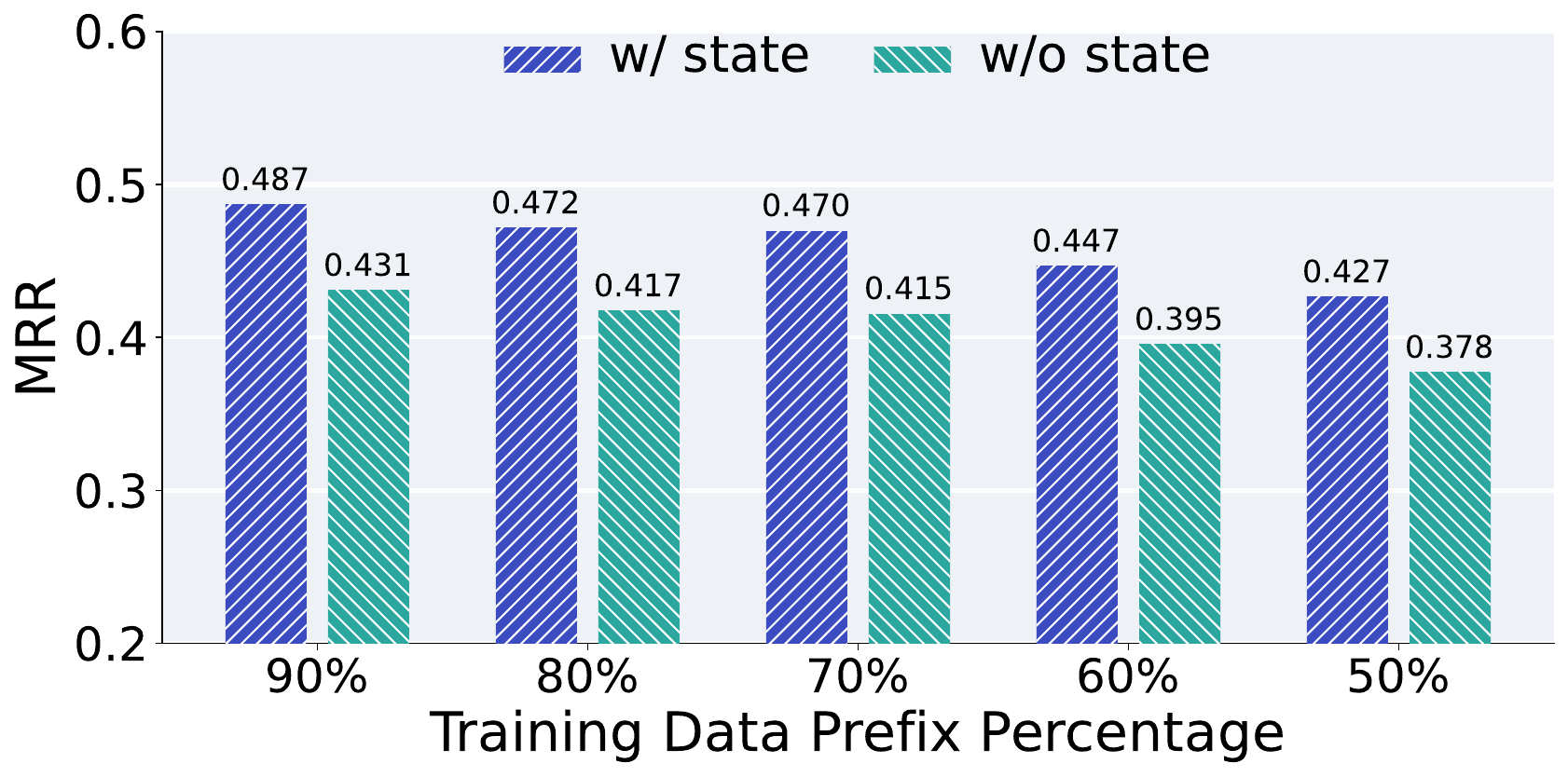}
    \caption{EST-Mamba on ICEWS14.}
    \label{fig:rq4_temporal_mamba_icews14}
  \end{subfigure}
  \hfill
  \begin{subfigure}[t]{0.48\linewidth}
    \centering
    \includegraphics[width=\linewidth]{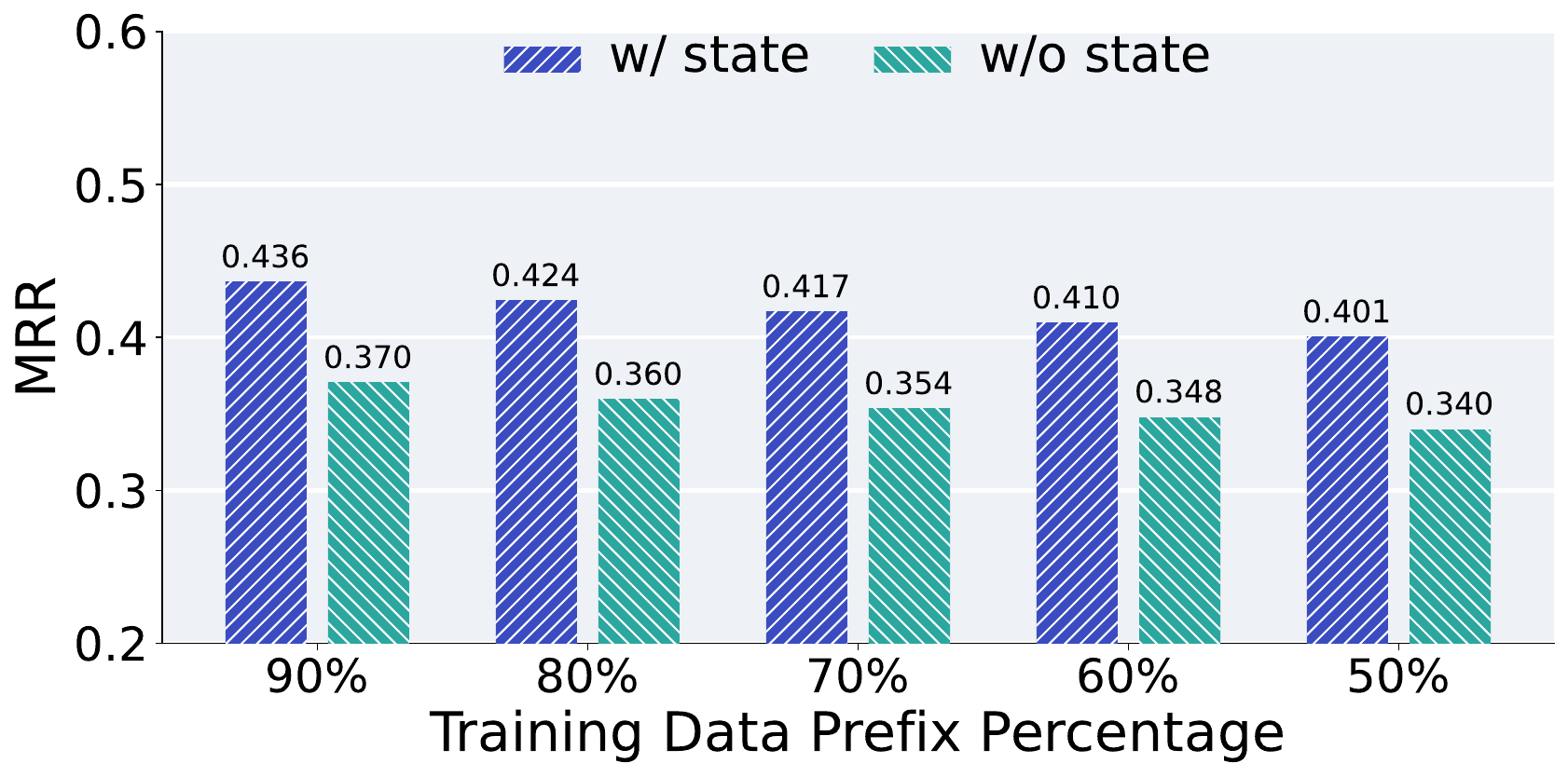}
    \caption{EST-Mamba on ICEWS18.}
    \label{fig:rq4_temporal_mamba_icews18}
  \end{subfigure}

  \vspace{-4pt}
  \caption{Temporal generalization under chronological data truncation. The first row shows EST-Transformer and the second row shows EST-Mamba; the left and right columns correspond to ICEWS14 and ICEWS18, respectively. EST consistently maintains a clear advantage over its stateless ablation across different training-history budgets.}
  \label{fig:rq4_temporal}
  \vspace{-3mm}
\end{figure*}

\subsection{Temporal Generalization: RQ4}
\label{sec:5.5}

To evaluate robustness under temporal distribution shift, we adopt a \textit{Chronological Data Truncation} protocol, where models are trained only on the earliest $X\%$ of the training history ($X \in \{50, 60, 70, 80, 90\}$).

Figure~\ref{fig:rq4_temporal} shows that EST consistently outperforms its stateless counterpart across all truncation levels on both ICEWS14 and ICEWS18, for both Transformer and Mamba backbones.
Although performance degrades as less history is observed, the advantage of persistent states remains stable, indicating that the benefit of EST is not tied to a particular sequence encoder but is intrinsic to the state-centric paradigm.

Notably, EST also exhibits strong data efficiency.
On ICEWS18 with only 50\% of the training history, EST-Transformer still achieves an MRR of 0.416, surpassing fully-supervised baselines such as DiffuTKG (0.367), while EST-Mamba reaches 0.401, also exceeding DiffuTKG and CognTKE (0.352).
These results suggest that persistent entity states capture transferable evolutionary dynamics beyond the observed window, leading to stronger temporal generalization than stateless modeling.

\section{Conclusion}

We identify a fundamental limitation in temporal knowledge graph forecasting: entities are commonly modeled as stateless across time, leading to repeated reconstruction and poor long-horizon generalization. To address this, we propose Entity State Tuning (EST), an encoder-agnostic framework that equips forecasters with persistent and continuously updated entity states. EST injects state priors into structural encoding, aggregates state-aware events with a pluggable temporal backbone, and updates entity states through an evolution mechanism that balances stability and plasticity, guided by a counterfactual objective. Experiments on four benchmarks show that EST consistently improves diverse backbones and achieves state-of-the-art results. Our findings highlight state persistence as central to robust TKG forecasting and suggest richer state evolution and memory consolidation beyond snapshot-based reasoning.

\section*{Limitations}
Notwithstanding EST's design as a universal and efficient paradigm, capturing the full spectrum of complex, non-stationary temporal dynamics remains a non-trivial challenge

First, it maintains global entity state buffers, introducing a memory overhead that scales linearly with the number of entities, i.e., $\mathcal{O}(|\mathcal{E}|d)$; while practical on standard benchmarks, this may become costly for web-scale graphs or memory-constrained deployments.
Second, EST assumes temporally ordered processing and a relatively stable entity universe, and thus does not explicitly address previously unseen entities or highly dynamic entity sets.
Finally, our evaluation is limited to four standard benchmarks under chronological splits; real-world TKGs may exhibit stronger non-stationarity, exogenous shocks, or richer supervision signals (e.g., attributes and text), which are beyond the scope of the current study.

\bibliography{custom}

\clearpage

\appendix

\crefalias{section}{appendix}
\crefalias{subsection}{appendix}
\crefalias{subsubsection}{appendix}
\crefalias{paragraph}{appendix}
\startcontents[appendix]
{
\hypersetup{linkcolor=black}
\printcontents[appendix]{}{0}{\section*{Appendix}}

\section{Datasets}
\label{app:datasets}

\begin{table*}[t]
\centering
\small
\caption{Statistics of Four Temporal Knowledge Graph Benchmarks}
\label{tab:tkg_datasets}
\begin{tabular}{lrrrrrr}
\toprule
\textbf{Dataset} & \textbf{\#Entities} & \textbf{\#Relations} & 
\textbf{\#Train} & \textbf{\#Valid} & \textbf{\#Test} & \textbf{\#Snapshots} \\
\midrule
ICEWS14    & 7,128  & 230 & 74,845    & 8,514   & 7,371   & 365  \\
ICEWS18    & 10,094 & 256 & 373,018   & 45,995  & 49,545  & 365  \\
ICEWS05-15 & 23,033 & 251 & 368,868   & 46,302  & 46,159  & 4,017 \\
GDELT      & 7,691  & 240 & 1,734,399 & 238,765 & 305,241 & 2,975 \\
\bottomrule
\end{tabular}
\end{table*}

\begin{table*}[h]
\centering
\small
\setlength{\aboverulesep}{0pt}
\setlength{\belowrulesep}{0pt}
\renewcommand{\arraystretch}{1.1}
\setlength{\tabcolsep}{6pt}

\caption{Impact of scoring functions on ICEWS14. }
\label{tab:scoring_fn_icews14}

\begin{subtable}[t]{0.48\linewidth}
\centering
\footnotesize
\begin{tabular}{l|cccc}
\toprule
\textbf{Scorer} & \textbf{MRR} & \textbf{Hit@1} & \textbf{Hit@3} & \textbf{Hit@10} \\
\midrule
DistMult & \textbf{0.511} & \textbf{0.451} & \textbf{0.538} & \textbf{0.633} \\
MLP      & \underline{0.485} & \underline{0.416} & \underline{0.516} & \underline{0.618} \\
ComplEx  & 0.439 & 0.371 & 0.466 & 0.570 \\
RotatE   & 0.471 & 0.399 & 0.504 & 0.609 \\
\bottomrule
\end{tabular}
\caption{EST-Mamba on ICEWS14.}
\label{tab:scoring_fn_mamba_icews14}
\end{subtable}
\hfill
\begin{subtable}[t]{0.48\linewidth}
\centering
\footnotesize
\begin{tabular}{l|cccc}
\toprule
\textbf{Scorer} & \textbf{MRR} & \textbf{Hit@1} & \textbf{Hit@3} & \textbf{Hit@10} \\
\midrule
DistMult & \textbf{0.513} & \textbf{0.455} & \textbf{0.536} & \textbf{0.629} \\
MLP      & \underline{0.475} & \underline{0.408} & \underline{0.502} & \underline{0.607} \\
ComplEx  & 0.434 & 0.366 & 0.460 & 0.566 \\
RotatE   & 0.470 & 0.400 & 0.501 & 0.604 \\
\bottomrule
\end{tabular}
\caption{EST-Transformer on ICEWS14.}
\label{tab:scoring_fn_tr_icews14}
\end{subtable}
\vspace{-4mm}
\end{table*}

We evaluate our approach on four widely used real-world temporal knowledge graph benchmarks: ICEWS14~\cite{garcia2018learning}, ICEWS18~\cite{jin2020recurrent}, ICEWS05-15~\cite{li2021temporal}, and GDELT~\cite{leetaru2013gdelt}. The three ICEWS datasets—ICEWS14, ICEWS18, and ICEWS05-15—are derived from the Integrated Crisis Early Warning System and consist of large-scale international political events spanning different temporal ranges. In contrast, GDELT is constructed from global news media and covers a broader spectrum of societal events across countries and regions.

For all datasets, we adopt the standard chronological splitting protocol with an 8:1:1 ratio for training, validation, and testing~\cite{jin2020recurrent}. To prevent temporal leakage, validation data are strictly drawn from timestamps following the training period, and test data are restricted to timestamps occurring after the validation window. Detailed dataset statistics, including the numbers of entities, relations, timestamps, and quadruples in each split, are reported in Table~\ref{tab:tkg_datasets}.

\section{Implementation Details}
\label{app:implementation_details}
\subsection{Experiment Settings}

Unless otherwise stated, all models are implemented in PyTorch and trained on a single NVIDIA GeForce RTX 4090 GPU.
We set the embedding dimension to $d=64$ and the time encoding dimension to $d_t=32$.
Notably, we employ a lightweight MLP as the default structural encoder $\Phi_{\text{struct}}$, ensuring that performance gains are attributed to the state tuning mechanism rather than complex topological aggregation.
For each query at time $\tau$, we retrieve an entity-centric history of length $L=32$ (strictly earlier than $\tau$).
Temporal gaps are encoded via a $\log(1+\Delta t)$ projection, and DistMult is adopted as the default scoring function.

To realize the proposed dual-system memory, we maintain two non-trainable entity state buffers, both of which are initialized to all-zero vectors for every entity at the beginning of training.
During training, the fast state is updated using an exponential moving average with update rate $\lambda=0.2$,
while the slow state is updated via a sigmoid-gated interpolation with fixed scale $\kappa=5.0$ and threshold $\gamma=0.5$. Overall, the end-to-end training and the closed-loop state evolution strictly follow \textbf{Algorithm~\ref{alg:est}}.

\subsection{Optimization.}
We train for 20 epochs with mini-batch size 128 using Adam with learning rate $5\times 10^{-3}$ and weight decay $10^{-4}$.
We adopt a linear warmup for the first 2 epochs and then apply cosine annealing scheduling, where the minimum learning rate is set to $0.1$ times the base learning rate.

\subsection{Evaluation protocol.}
We report Mean Reciprocal Rank (MRR) and Hits@K ( K $\in$ \{1,3,10\}) under a time-consistent filtered ranking protocol, analogous to the standard filtered evaluation setting commonly adopted in TKG.
Specifically, we evaluate queries in chronological order and only filter candidate tails that were observed strictly before the current timestamp (rolling setting), then reveal events at the current timestamp after evaluation.
This avoids temporal leakage in filtered ranking and ensures  counterfactual consistency.

\section{Scoring Function}
\label{app:scoring_function}
We instantiate the decoder with four scoring functions (DistMult, RotatE, ComplEx, and an MLP).
On ICEWS14, DistMult consistently achieves the best results for both EST-Mamba (MRR 0.511, Hits@1 0.451, Hits@3 0.538, Hits@10 0.633) and EST-Transformer (MRR 0.513, Hits@1 0.455, Hits@3 0.536, Hits@10 0.629).
This suggests that the state-enhanced temporal context produced by EST can be effectively decoded by a lightweight bilinear scorer, while more expressive complex/rotational decoders do not yield additional benefits.
We therefore adopt DistMult as the default scorer in all experiments.

\section{Sequential Backbones}
\label{app:backbones}

As outlined in Section~\ref{sec:4.2}, the EST framework is designed to be encoder-agnostic, abstracting the temporal evolution into a generalized state transition function $\mathcal{T}$.
Formally, given the sequence of structure-aware event representations $\mathbf{U} = [\mathbf{u}_1, \ldots, \mathbf{u}_L] \in \mathbb{R}^{L \times d}$ derived from the Unified Temporal Context Module, the sequential encoder maps $\mathbf{U}$ to a sequence of contextualized hidden states $\mathbf{H} = [\mathbf{h}_1, \ldots, \mathbf{h}_L]$ under the constraint of temporal visibility (i.e., $\mathbf{h}_t$ depends only on $\mathbf{u}_{\leq t}$).

In this section, we provide the specific mathematical formulations for the four instantiated backbones: RNN~\cite{sherstinsky2020fundamentals}, LSTM~\cite{hochreiter1997long}, Transformer~\cite{vaswani2017attention}, and Mamba~\cite{gu2024mamba}. Table~\ref{tab:seq_math_analysis} presents a comparative analysis of their update dynamics, receptive fields, time complexity, and optimization properties.

\newcolumntype{P}[1]{>{\raggedright\arraybackslash\hspace{0pt}}p{#1}}

\begin{table*}[t]
\centering
\footnotesize
\setlength{\tabcolsep}{4pt}
\renewcommand{\arraystretch}{1.2}
\caption{Comparative analysis of sequential backbone instantiations within the EST framework.
We contrast the core mathematical form, temporal mixing mechanism, theoretical receptive field, and computational complexity for each backbone. $L$ denotes the sequence length, and $d$ represents the hidden dimension.}
\label{tab:seq_math_analysis}

\resizebox{\textwidth}{!}{
\begin{tabular}{P{1.8cm} P{1.8cm} P{4.5cm} P{4.5cm} P{2.4cm} P{3.6cm}}
\toprule
\textbf{Backbone} &
\textbf{Core Mechanism} &
\textbf{Temporal Mixing Operator $\mathcal{T}$} &
\textbf{Receptive Field \& Dependencies} &
\textbf{Time Complexity (Train/Infer)} &
\textbf{Gradient Flow \& Stability} \\
\midrule

\textbf{RNN} &
Non-linear Recurrence &
\parbox[t]{\linewidth}{\vspace{2pt}%
\(\displaystyle \mathbf{h}_t = \tanh(\mathbf{W}_u \mathbf{u}_t + \mathbf{W}_h \mathbf{h}_{t-1})\)\\
\emph{(Implicit 1-step Markov mixing)}%
\vspace{2pt}} &
\parbox[t]{\linewidth}{\vspace{2pt}%
Infinite impulse response (theoretically); implies dependency on full history $\mathbf{u}_{<t}$ via recursion.%
\vspace{2pt}} &
\parbox[t]{\linewidth}{\vspace{2pt}%
\(\mathcal{O}(Ld^2)\)\\
\(\mathcal{O}(Ld^2)\)%
\vspace{2pt}} &
\parbox[t]{\linewidth}{\vspace{2pt}%
Susceptible to vanishing/exploding gradients over long $L$; lacks explicit gating stabilization.%
\vspace{2pt}}
\\
\midrule

\textbf{LSTM} &
Gated Recurrence &
\parbox[t]{\linewidth}{\vspace{2pt}%
\(\displaystyle \mathbf{c}_t = \mathbf{f}_t \odot \mathbf{c}_{t-1} + \mathbf{i}_t \odot \tilde{\mathbf{c}}_t\)\\
\(\displaystyle \mathbf{h}_t = \mathbf{o}_t \odot \tanh(\mathbf{c}_t)\)%
\vspace{2pt}} &
\parbox[t]{\linewidth}{\vspace{2pt}%
Causal; the additive memory cell $\mathbf{c}_t$ creates a "constant error carousel" to retain long-term dependencies.%
\vspace{2pt}} &
\parbox[t]{\linewidth}{\vspace{2pt}%
\(\mathcal{O}(Ld^2)\)\\
\(\mathcal{O}(Ld^2)\)%
\vspace{2pt}} &
\parbox[t]{\linewidth}{\vspace{2pt}%
Gating mechanisms alleviate gradient decay; stable optimization for medium-to-long sequences.%
\vspace{2pt}}
\\
\midrule

\textbf{Transformer} &
Global Attention &
\parbox[t]{\linewidth}{\vspace{2pt}%
\(\displaystyle \mathbf{A} = \mathrm{softmax}\left(\frac{\mathbf{Q}\mathbf{K}^\top}{\sqrt{d}} + \mathbf{M}\right)\mathbf{V} \)\\
\emph{(Content-based global kernel)}%
\vspace{2pt}} &
\parbox[t]{\linewidth}{\vspace{2pt}%
Global receptive field ($L$); direct path between any pair $(t, \tau)$ within the window regardless of distance.%
\vspace{2pt}} &
\parbox[t]{\linewidth}{\vspace{2pt}%
\(\mathcal{O}(L^2 d)\)\\
\(\mathcal{O}(L^2 d)\)%
\vspace{2pt}} &
\parbox[t]{\linewidth}{\vspace{2pt}%
Shortest gradient path ($\mathcal{O}(1)$); highly parallelizable but memory-intensive for large $L$.%
\vspace{2pt}}
\\
\midrule

\textbf{Mamba} &
Selective State Space &
\parbox[t]{\linewidth}{\vspace{2pt}%
\(\displaystyle \mathbf{s}_t = \bar{\mathbf{A}}_t \mathbf{s}_{t-1} + \bar{\mathbf{B}}_t \mathbf{u}_t\)\\
\(\displaystyle \mathbf{h}_t = \mathbf{C}_t \mathbf{s}_t\)\\
\emph{(Input-dependent Linear Scan)}%
\vspace{2pt}} &
\parbox[t]{\linewidth}{\vspace{2pt}%
Causal; combines RNN-like recursion with CNN-like parallel scan. Selection mechanism compresses context adaptively.%
\vspace{2pt}} &
\parbox[t]{\linewidth}{\vspace{2pt}%
\(\mathcal{O}(L d)\)\\
\(\mathcal{O}(L d)\)%
\vspace{2pt}} &
\parbox[t]{\linewidth}{\vspace{2pt}%
Linear scalability with stable long-range propagation due to discretized state evolution.%
\vspace{2pt}}
\\
\bottomrule
\end{tabular}
}
\end{table*}

\begin{table*}[t]
\centering
\caption{Worked example of history retrieval and query-specific attention on ICEWS14. For brevity, we show the most attended events in the retrieved history window.}
\renewcommand{\arraystretch}{1.05}
\setlength{\tabcolsep}{12pt}
\label{tab:worked_example}
\begin{tabular}{c|c|c|c}
\toprule
Order & Historical event & $\Delta t$ & Attention \\
\midrule
1 & (\texttt{Refugee\_(Afghanistan)}, \texttt{Make\_a\_visit}, \texttt{Iran}, 7944) & 72   & 0.1753 \\
2 & (\texttt{Refugee\_(Afghanistan)}, \texttt{Make\_a\_visit}, \texttt{Iran}, 4728) & 3288 & 0.1403 \\
3 & (\texttt{Refugee\_(Afghanistan)}, \texttt{Make\_a\_visit}, \texttt{Iran}, 5304) & 2712 & 0.1412 \\
4 & (\texttt{Refugee\_(Afghanistan)}, \texttt{Make\_a\_visit}, \texttt{Iran}, 4848) & 3168 & 0.1380 \\
5 & (\texttt{Refugee\_(Afghanistan)}, \texttt{Make\_a\_visit}, \texttt{Iran}, 4992) & 3024 & 0.1344 \\
\bottomrule
\end{tabular}
\end{table*}

As illustrated in Table~\ref{tab:seq_math_analysis}, each backbone offers distinct trade-offs. RNN and LSTM offer linear complexity with respect to sequence length $L$ but suffer from sequential computation bottlenecks and gradient decay.
Transformers provide the strongest direct modeling of long-term dependencies via attention, yet their quadratic complexity $\mathcal{O}(L^2)$ hinders scalability for extremely long histories.
Mamba represents a compelling middle ground, achieving the linear scaling of RNN ($\mathcal{O}(L)$) while approximating the expressive power of Transformers through input-dependent selection mechanisms.
By supporting these diverse architectures, EST allows practitioners to tailor the balance between computational efficiency and modeling capacity based on specific deployment constraints.

\section{End-to-End Workflow on ICEWS14}
\label{app:worked_example}

To make the formalization in Sections~\ref{sec.3}--\ref{sec.4} more concrete, we present a worked example from ICEWS14 that instantiates Algorithm~\ref{alg:est} on a real test query.

\textbf{Step 1. Query formulation.}
Consider the test query: 
\[
\small{
(\texttt{Refugee\_(Afghanistan)}, \texttt{Make\_a\_visit}, ?, 8016),}
\]
whose gold tail is \texttt{Iran}. EST treats this as a temporal forecasting query and retrieves the history of the head entity before timestamp $t=8016$.

\textbf{Step 2. History retrieval and attention.}
Table~\ref{tab:worked_example} lists the most attended events in the retrieved history window. The history shows a repeated interaction pattern with \texttt{Iran} under the relation \texttt{Make\_a\_visit}, and the most recent event receives the largest attention weight, consistent with temporal forecasting intuition.

\textbf{Step 3. Forward scoring.}
Each retrieved event is encoded into a state-aware structural representation and then passed through the sequence backbone to produce contextualized hidden states. Query-specific attention aggregates them into a context vector $\mathbf{c}_s$, which is used to score candidate tails via $\phi(\mathbf{c}_s, r, o)$. In this example, repeated visits to \texttt{Iran} provide the dominant evidence for ranking the gold tail.

\textbf{Step 4. Persistent-state write-back.}
After the forward pass, EST writes the resulting context back to the persistent states of the head entity through the fast and slow buffers. This closed-loop update allows future predictions involving \texttt{Refugee\_(Afghanistan)} to depend on accumulated temporal evidence rather than being reconstructed independently at each snapshot.

\textbf{Takeaway.}
This example illustrates how EST processes a real query end-to-end: query formulation, history retrieval, query-aware aggregation, tail scoring, and persistent state evolution.

\begin{algorithm*}[t]
\caption{Entity State Tuning (EST) for Temporal KG Extrapolation}
\label{alg:est}
\DontPrintSemicolon
\KwIn{Training set $\mathcal{D}$; entity/relation embeddings $\mathbf{E}_{\mathcal{E}},\mathbf{E}_{\mathcal{R}}$; global states $(\mathbf{S}_{\text{fast}},\mathbf{S}_{\text{slow}})$; structural encoder $\Phi_{\text{struct}}$; sequence backbone $\mathcal{T}$; time map $\varphi(\cdot)$; hyperparams $\lambda,\kappa,\gamma$; negative sampler $\mathrm{Neg}(\cdot)$.}
\KwOut{Model parameters $\Theta$ and updated global states $(\mathbf{S}_{\text{fast}},\mathbf{S}_{\text{slow}})$.}

\BlankLine
\For{each minibatch $\mathcal{B}\subset\mathcal{D}$}{
    $\mathcal{L}\leftarrow 0$\;
    \For{each quadruple $(s,r,o,t)\in\mathcal{B}$}{
        Construct history $\mathcal{H}_s(t)=\{(o_k,r_k,\tau_k)\}_{k=1}^{L}$\;

        \tcp{(1) Topology-Aware State Perceiver}
        \For{$k=1$ \KwTo $L$}{
            $\mathbf{S}_{o_k}\leftarrow \mathbf{S}_{\text{slow}}(o_k)$\;
            $\mathbf{g}_k \leftarrow \sigma\!\left(\mathbf{W}_g[\mathbf{E}_{o_k}\,\|\,\mathbf{S}_{o_k}]\right)$\;
            $\tilde{\mathbf{E}}_{o_k}\leftarrow \mathbf{g}_k\odot\mathbf{E}_{o_k} + (1-\mathbf{g}_k)\odot\mathbf{S}_{o_k}$\;
            $\mathbf{X}_k \leftarrow \Phi_{\text{struct}}\!\left(\tilde{\mathbf{E}}_{o_k},\mathbf{E}_{r_k}\right)$\;
            $\mathbf{u}_k \leftarrow \mathrm{MLP}\!\left([\mathbf{X}_k \,\|\, \mathbf{E}_{r_k} \,\|\, \varphi(t-\tau_k)]\right)$\;
        }

        \tcp{(2) Unified Temporal Context Module}
        $\mathbf{h}_{1:L} \leftarrow \mathcal{T}(\mathbf{u}_{1:L})$, \quad $\mathbf{y}_k \leftarrow \mathcal{P}(\mathbf{h}_k)$\;
        Compute query-aware attention $\{\alpha_k\}_{k=1}^{L}$ and context $\mathbf{c}_s \leftarrow \sum_{k=1}^{L}\alpha_k\mathbf{y}_k$\;

        \tcp{(3) De-confounded State Evolution}
        $\mathcal{E}^- \leftarrow \mathrm{Neg}(s,r,t)$\;
        $\mathcal{L}\leftarrow \mathcal{L} - \phi(\mathbf{c}_s,r,o) + \log\!\sum\limits_{o'\in\{o\}\cup\mathcal{E}^-}\exp\!\left(\phi(\mathbf{c}_s,r,o')\right)$\;

        \tcp{Closed-loop state evolution (dual-system update)}
        $\mathbf{S}_{\text{fast}}(s)\leftarrow (1-\lambda)\mathbf{S}_{\text{fast}}(s) + \lambda\,\mathbf{c}_s$\;
        $\delta_s \leftarrow \|\mathbf{S}_{\text{fast}}(s)-\mathbf{S}_{\text{slow}}(s)\|_2$\;
        $g_s \leftarrow \sigma\!\big(\kappa(\delta_s-\gamma)\big)$\;
        $\mathbf{S}_{\text{slow}}(s)\leftarrow \mathbf{S}_{\text{slow}}(s) + g_s\big(\mathbf{S}_{\text{fast}}(s)-\mathbf{S}_{\text{slow}}(s)\big)$\;
    }
    Update $\Theta$ by minimizing $\mathcal{L}$\;
}
\end{algorithm*}

\end{document}